\documentclass[10pt,twocolumn,letterpaper]{article}

\usepackage{cvpr}
\usepackage{times}
\usepackage{epsfig}
\usepackage{graphicx}
\usepackage{amsmath}
\usepackage{amssymb}
\usepackage{amsmath,amsthm,amsfonts}

\usepackage{comment}
\usepackage{algorithm}
\usepackage{algorithmic}

\usepackage{graphicx}
\usepackage{subfigure}
\usepackage{multirow}
\usepackage{enumerate}
\usepackage{pifont}
\usepackage{array}
\usepackage{booktabs}
\usepackage[T1]{fontenc}

\usepackage[
  separate-uncertainty = true,
  multi-part-units = repeat
]{siunitx}
\newcommand{\tabincell}[2]{\begin{tabular}{@{}#1@{}}#2\end{tabular}}
\newcommand{\topcaption}{%
\setlength{\abovecaptionskip}{0pt}%
\setlength{\belowcaptionskip}{5pt}%
\caption}
\usepackage[breaklinks=true,bookmarks=false]{hyperref}

\cvprfinalcopy % *** Uncomment this line for the final submission

\def\cvprPaperID{8050} % *** Enter the CVPR Paper ID here

\ifcvprfinal\fi
\begin{document}

%%%%%%%%% TITLE
\title{Combating Noisy Labels by Agreement: \\ A Joint Training Method with Co-Regularization}

\author{Hongxin Wei$^{1}$ \quad Lei Feng$^{1}$\thanks{Corresponding author.} \quad  Xiangyu Chen$^{2}$ \quad  Bo An$^{1}$\\
$^1$School of Computer Science and Engineering, Nanyang Technological University, Singapore \quad\\
$^2$Open FIESTA Center, Tsinghua University, China\\
{\tt\small \{owenwei,boan\}@ntu.edu.sg \quad feng0093@e.ntu.edu.sg \quad chenxian18@mails.tsinghua.edu.cn}
}

\maketitle
\thispagestyle{empty}

\begin{abstract}
Deep Learning with noisy labels is a practically challenging problem in weakly supervised learning. The state-of-the-art approaches ``Decoupling" and ``Co-teaching+" claim that the ``disagreement" strategy is crucial for alleviating the problem of learning with noisy labels. In this paper, we start from a different perspective and propose a robust learning paradigm called JoCoR, which aims to reduce the diversity of two networks during training. Specifically, we first use two networks to make predictions on the same mini-batch data and calculate a joint loss with Co-Regularization for each training example. Then we select small-loss examples to update the parameters of both two networks simultaneously. Trained by the joint loss, these two networks would be more and more similar due to the effect of Co-Regularization. Extensive experimental results on corrupted data from benchmark datasets including MNIST, CIFAR-10, CIFAR-100 and Clothing1M demonstrate that JoCoR is superior to many state-of-the-art approaches for learning with noisy labels.

\vspace{-10pt}

\end{abstract}
\section{Introduction}

Deep Neural Networks (DNNs) achieve remarkable success on various tasks, and most of them are trained in a supervised manner, which heavily relies on a large number of training instances with accurate labels  \cite{he2016deep}. 
However, collecting large-scale datasets with fully precise annotations is expensive and time-consuming. To alleviate this problem, data annotation companies choose some alternating methods such as crowdsourcing \cite{yan2014learning,yu2018learning} and online queries \cite{blum2003noise} to improve labelling efficiency. 
Unfortunately, these methods usually suffer from unavoidable noisy labels, which have been proven to lead to noticeable decrease in performance of DNNs \cite{arpit2017closer,zhang2016understanding}.

As this problem has severely limited the expansion of neural network applications, a large number of algorithms have been developed for learning with noisy labels, which belongs to the family of weakly supervised learning frameworks \cite{Bao2018Classification,Chapelle2006semi,Plessis2014Analysis,feng2018leveraging,feng2019partial2,feng2019partial,gong2017learning}. Some of them focus on improving the methods to estimate the latent noisy transition matrix \cite{liu2015classification,menon2015learning,sanderson2014class}. However, it is challenging to estimate the noise transition matrix accurately. An alternative approach is training on selected or weighted samples, e.g., Mentornet \cite{jiang2017mentornet}, gradient-based reweight \cite{ren2018learning} and Co-teaching \cite{han2018co}. 
% while others work on deep learning solutions to deal with noisy labels \cite{wang2018iterative,veit2017learning,li2017learning}.
 % Among them, ``small-loss" criterion has been widely applied to select clean examples for reducing the negative effect of noisy labels. 
 Furthermore, the state-of-the-art methods including Co-teaching+ \cite{yu2019does} and Decoupling \cite{malach2017decoupling} have shown excellent performance in learning with noisy labels by introducing the ``Disagreement" strategy, where ``when to update" depends on a disagreement between two different networks. However, there are only a part of training examples that can be selected by the ``Disagreement" strategy, and these examples cannot be guaranteed to have ground-truth labels \cite{han2018co}. Therefore, there arises a question to be answered: Is ``Disagreement" necessary for training two networks to deal with noisy labels?

Motivated by Co-training for multi-view learning and semi-supervised learning that aims to maximize the agreement on multiple distinct views \cite{blum1998combining,kumar2010co,sindhwani2005co,zhang2018deep}, a straightforward method for handling noisy labels is to apply the regularization from peer networks when training each single network. However, although the regularization may improve the generalization ability of networks by encouraging agreement between them, it still suffers from memorization effects on noisy labels \cite{zhang2016understanding}. To address this problem, we propose a novel approach named JoCoR (\textbf{Jo}int Training with \textbf{Co-R}egularization). Specifically, we train two networks with a joint loss, including the conventional supervised loss and the Co-Regularization loss. Furthermore, we use the joint loss to select small-loss examples, thereby ensuring the error flow from the biased selection would not be accumulated in a single network.

To show that JoCoR significantly improves the robustness of deep learning on noisy labels, we conduct extensive experiments on both simulated and real-world
noisy datasets, including MNIST, CIFAR-10, CIFAR-100 and Clothing1M datasets. Empirical results demonstrate that the robustness of deep models trained by our proposed approach is superior to many state-of-the-art approaches. Furthermore, the ablation studies clearly demonstrate the effectiveness of Co-Regularization and Joint Training.

\section{Related work}
In this section, we briefly review existing works on learning with noisy labels.  %Generally speaking, the existing works can be 

\textbf{Noise rate estimation}. The early methods focus on estimating the label transition matrix \cite{menon2015learning,natarajan2013learning,patrini2017making,xia2019anchor}. 
For example, F-correction \cite{patrini2017making} uses a two-step solution to heuristically estimate the noise transition matrix. An additional softmax layer is introduced to model the noise transition matrix \cite{goldberger2016training}. 
In these approaches, the quality of noise rate estimation is a critical factor for improving robustness. 
However, noise rate estimation is challenging, especially on datasets with a large number of classes.

\textbf{Small-loss selection}. Recently, a promising method of handling noisy labels is to train models on small-loss instances \cite{ren2018learning}. Intuitively, the performance of DNNs will be better if the training data become less noisy. Previous work showed that during training, DNNs tend to learn simple patterns first, then gradually memorize all samples \cite{arpit2017closer}, which justifies the widely used \textit{small-loss criterion}: treating samples with small training loss as clean ones. In particular, MentorNet \cite{jiang2017mentornet} firstly trains a teacher network, then uses it to select clean instances for guiding the training of the student network. As for Co-teaching \cite{han2018co}, in each mini-batch of data, each network chooses its small-loss instances and exchanges them with its peer network for updating the parameters. The authors argued that these two networks could filter different types of errors brought by noisy labels since they have different learning abilities. When the error from noisy data flows into the peer network, it will attenuate this error due to its robustness.

\begin{figure}[!t]
\centering
\includegraphics[scale=0.39]{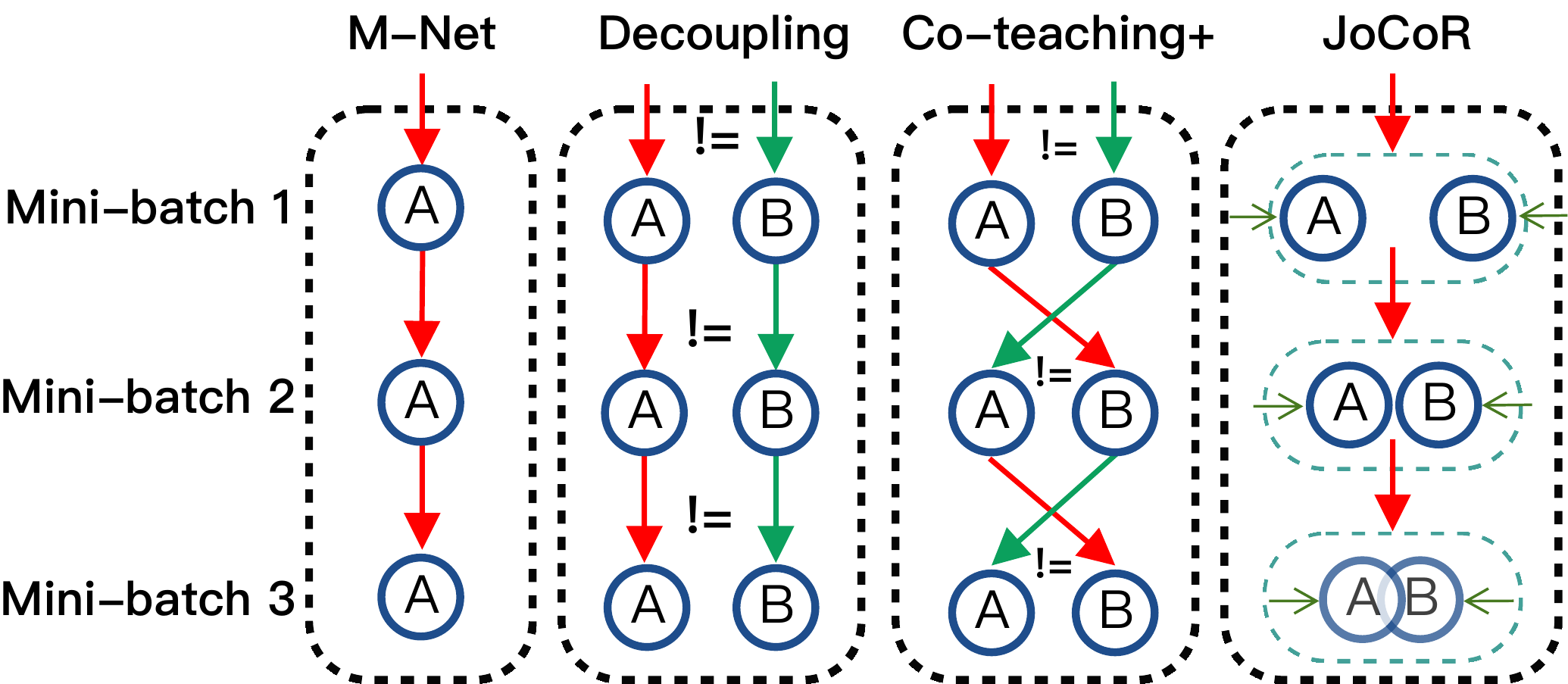}
\caption{Comparison of error flow among MentorNet (M-Net) \cite{jiang2017mentornet}, Decoupling \cite{malach2017decoupling}, Co-teaching+ \cite{yu2019does} and JoCoR. Assume that the error flow comes from the biased selection of training instances, and error flow from network A or B is denoted by red arrows or green arrows, respectively. \textbf{First panel}: M-Net maintains only one network (A). \textbf{Second panel}: Decoupling maintains two networks (A\&B). The parameters of two networks are updated, when the predictions of them disagree (!=). \textbf{Third panel}: In Co-teaching+, each network teaches its small-loss instances with prediction disagreement (!=) to its peer network. \textbf{Fourth panel}: JoCoR also maintains two networks (A\&B) but trains them as a whole with a joint loss, which makes predictions of each network closer to ground true labels and peer network's.
%In each mini-batch data, two networks feed forward and make predictions for each instance, then select small-loss instances based on the joint loss for further training.
}
\label{fig:error-flow}
\vspace{-18pt}
\end{figure}

\textbf{Disagreement}. The ``Disagreement" strategy is also applied to this problem. For instance, Decoupling \cite{malach2017decoupling} updates the model only using instances on which the predictions of two different networks are different. The idea of disagreement-update is similar to hard example mining \cite{shrivastava2016training}, which trains model with examples that are misclassified and expects these examples to help steer the classifier away from its current mistakes. For the ``Disagreement" strategy, the decision of ``when to update" depends on a disagreement between two networks instead of depending on the label. As a result, it would help decrease the divergence between these networks.
%Intuitively, the idea of disagreement-update also comes from Co-training \cite{blum1998combining}, where two classifiers should keep diverged in order to achieve a better ensemble effect.%
However, as noisy labels are spread across the whole space of examples, there may be many noisy labels in the disagreement area, where the Decoupling approach cannot handle noisy labels explicitly. Combining the ``Disagreement" strategy with cross-update in Co-teaching, Co-teaching+ \cite{yu2019does} achieves excellent performance in improving the robustness of DNNs against noisy labels. In spite of that, Co-teaching+ only selects small-loss instances with different predictions from two models so very few examples are utilized for training in each mini-batch when datasets are with extremely high noise rate. It would prevent the training process from efficient use of training examples. This phenomenon will be explicitly shown in our experiments in the symmetric-80\% label noise case.

\textbf{Other deep learning methods}. In addition to the aforementioned approaches, there are some other deep learning solutions \cite{han2019deep,kim2019nlnl} to deal with noisy labels, including pseudo-label based \cite{tanaka2018joint,yi2019probabilistic} and robust loss based approaches \cite{patrini2017making,zhang2018generalized}. For pseudo-label based approaches, Joint optimization \cite{tanaka2018joint} learns network parameters and infers the ground-true labels simultaneously. PENCIL \cite{yi2019probabilistic} adopts label probability distributions to supervise network learning and to update these distributions through back-propagation end-to-end in each epoch. For robust loss based approaches, F-correct\cite{patrini2017making} proposes a robust risk minimization method to learn neural networks for multi-class classification by estimating label corruption probabilities. GCE \cite{zhang2018generalized} combines the advantages of the mean absolute loss and the cross entropy loss to obtain a better loss function and presents a theoretical analysis of the proposed loss functions in the context of noisy labels.

\textbf{Semi-supervised learning}. Semi-supervised learning also belongs to the family of weakly supervised learning frameworks \cite{Ishida2018Binary,kiryo2017positive,lu2019on, Niu2016Theoretic,niu2013squared,sakai2017semi,zhou2018brief}. There are some interesting works from semi-supervised learning that are highly relevant to our approach. In contrast to ``Disagreement" strategy, many of them are based on a agreement maximization algorithm. Co-RLS \cite{sindhwani2005co} extends standard regularization methods like Support Vector Machines (SVM) and Regularized Least squares (RLS) to multi-view semi-supervised learning by optimizing measures of agreement and smoothness over labelled and unlabelled examples. EA++ \cite{kumar2010co} is a co-regularization based approach for semi-supervised domain adaptation, which builds on the notion of augmented space and harnesses unlabeled data in the target domain to further assist the transfer of information from source to target. The intuition is that different models in each view would agree on the labels of most examples, and it is unlikely for compatible classifiers trained on independent views to agree on an incorrect label. This intuition also motivates us to deal with noisy labels based on the agreement maximization principle.

% A related idea from model distillation is Deep Mutual learning (DML)\cite{zhang2018deep}, where each student is asked to match the predictions of the ensemble of all other students in the cohort. With this approach, networks could perform better than those distilled from a strong but static teacher. The reason is that regularization from peer networks helps the model to find a much wider minima, which is expected to provide better generalization performance. This phenomenon can also explain why our algorithm could achieve the best test accuracy even when its label precision is slightly lower than Co-teaching. However, the only similarity between JoCoR and DML is that both of them reduce the disagreement between two classifiers. Specifically, DML trains two classifiers separately, while JoCoR builds a joint loss to update their parameters together following Co-RLS\cite{sindhwani2005co}. Additionally, DML mainly applies in the area of \textit{model distillation} and can be extended to \textit{semi-supervised learning}(SSL), but JoCoR is designed for \textit{learning with noisy labels}(LNL). As LNL is not a special case of SSL, we cannot simply translate DML from one problem setting to another problem setting.

\section{The Proposed Approach}

As mentioned before, we suggest to apply the agreement maximization principle to tackle the problem of noisy labels. In our approach, we encourage two different classifiers to make predictions closer to each other by explicit regularization method instead of hard sampling employed by the “Disagreement” strategy. This method could be considered as a meta-algorithm that trains two base classifiers by one loss function, which includes a regularization term to reduce divergence between the two classifiers.

For multi-class classification with $M$ classes, we suppose the dataset with $N$ samples is given as $D = \{\boldsymbol{x}_i,y_i\}_{i=1}^N$, where $\boldsymbol{x}_i$ is the $i$-th instance with its observed label as $y_i \in \{1,\ldots,M\}$. Similar to Decoupling and Co-teaching+, we formulate the proposed JoCoR approach with two deep neural networks denoted by $f(\boldsymbol{x},\boldsymbol{\Theta}_1)$ and $f(\boldsymbol{x},\boldsymbol{\Theta}_2)$, while $\boldsymbol{p}_1 = [p_1^1, p_1^2,\ldots, p_1^M]$ and $\boldsymbol{p}_2 = [p_2^1, p_2^2,\ldots, p_2^M]$ denote their prediction probabilities of instance $\boldsymbol{x}_i$, respectively. In other words, $\boldsymbol{p}_1$ and $\boldsymbol{p}_2$ are the outputs of the ``softmax" layer in $\boldsymbol{\Theta}_1$ and $\boldsymbol{\Theta}_2$.

\begin{figure}[!t]
\centering
\includegraphics[scale=0.45]{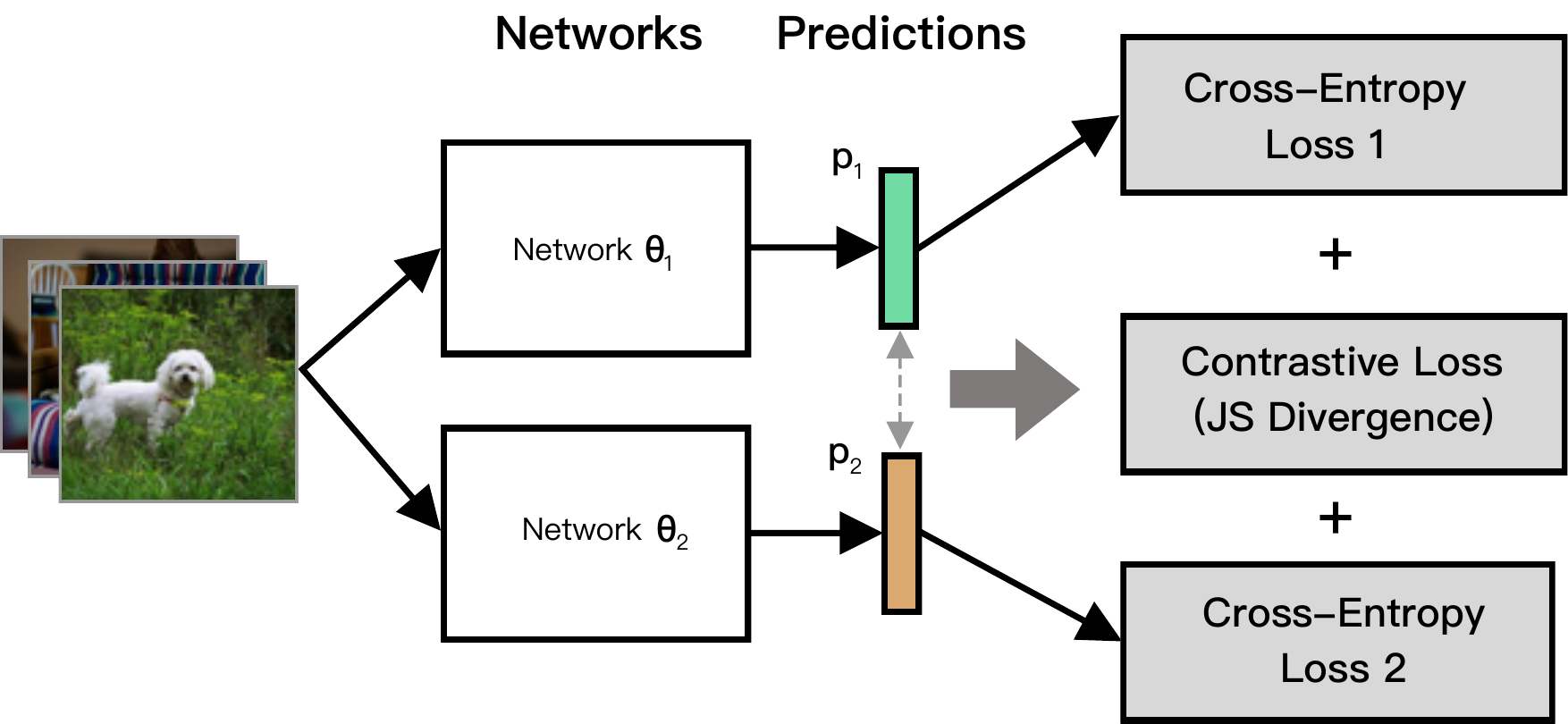}
\caption{JoCoR schematic.}
\label{fig:network}
\vspace{-10pt}
\end{figure}

\noindent\textbf{Network}. For JoCoR, each network can be used to predict labels alone, but during the training stage these two networks are trained with a pseudo-siamese paradigm, which means their parameters are different but updated simultaneously by a joint loss (see Figure \ref{fig:network}). In this work, we call this paradigm as ``Joint Training".

Specifically, our proposed loss function $\ell$ on $\boldsymbol{x}_i$ is constructed as follows:
\begin{equation}
\label{eq:joint_loss}
\ell(x_i) = (1-\lambda) * \ell_{\text{sup}}(\boldsymbol{x}_i,y_i) + \lambda * \ell_{\text{con}}(\boldsymbol{x}_i)
\end{equation}
In the loss function, the first part $\ell_{\text{sup}}$ is conventional supervised learning loss of the two networks, the second part $\ell_{\text{con}}$ is the contrastive loss between predictions of the two networks for achieving Co-Regularization.

\noindent\textbf{Classification loss}. For multi-class classification, we use Cross-Entropy Loss as the supervised part to minimize the distance between predictions and labels. 
\vspace{-5pt}
\begin{equation}
\begin{aligned}
\ell_{\text{sup}}(\boldsymbol{x}_i,y_i) &= \ell_{\text{C1}}(\boldsymbol{x}_i,y_i) + \ell_{\text{C2}}(\boldsymbol{x}_i,y_i)\\
& = -\sum\nolimits_{i=1}^{N}\sum\nolimits_{m=1}^{M} y_i\log(p_1^m(\boldsymbol{x}_i)) \\
& -\sum\nolimits_{i=1}^{N}\sum\nolimits_{m=1}^{M} y_i\log(p_2^m(\boldsymbol{x}_i))
\end{aligned}
\vspace{-5pt}
\end{equation}

% with $I(y_i,m) = \begin{cases} 1& y_i=m\\ 0& y_i\neq m \end{cases}$

Intuitively, two networks can filter different types of errors brought by noisy labels since they have different learning abilities. In Co-teaching \cite{han2018co}, when the two networks exchange the selected small-loss instances in each mini-batch data, the error flows can be reduced by peer networks mutually. By virtue of the joint-training paradigm, our JoCoR would consider the classification losses from both two networks during the ``small-loss" selection stage. In this way, JoCoR can share the same advantage of the cross-update strategy in Co-teaching. This argument will be clearly supported by the ablation study in the later section.

\noindent\textbf{Contrastive loss}. From the view of \textit{agreement maximization principles} \cite{blum1998combining,sindhwani2005co}, different models would agree on labels of most examples, and they are unlikely to agree on incorrect labels.
% the agreed examples are more likely to have correct labels. 
Based on this observation, we apply the Co-Regularization method to maximize the agreement between two classifiers. On one hand, the Co-Regularization term could help our algorithm select examples with clean labels since an example with small Co-Regularization loss means that two networks reach an agreement on its predictions. On the other hand, the regularization from peer networks helps the model find a much wider minimum, which is expected to provide better generalization performance \cite{zhang2018deep}. 

 In JoCoR, we utilize the contrastive term as Co-Regularization to make the networks guide each other. To measure the match of the two networks' predictions $\boldsymbol{p}_1$ and $\boldsymbol{p}_2$, we adopt the Jensen-Shannon (JS) Divergence. To simplify implementation, we could use the symmetric Kullback-Leibler(KL) Divergence to surrogate this term.
 \vspace{-5pt}
\begin{equation}
\begin{aligned}
\label{eq:contrastive}
\ell_{\text{con}} = D_{\text{KL}}(\boldsymbol{p}_1 || \boldsymbol{p}_2) + D_{\text{KL}}(\boldsymbol{p}_2 || \boldsymbol{p}_1)
\end{aligned}
\end{equation}
\noindent where
\begin{equation}
\nonumber
\begin{aligned}
\label{eq:KL}
D_{\text{KL}}(\boldsymbol{p}_1 || \boldsymbol{p}_2) = \sum\nolimits_{i=1}^{N}\sum\nolimits_{m=1}^{M}p_1^m(\boldsymbol{x}_i)\log\frac{p_1^m(\boldsymbol{x}_i)}{p_2^m(\boldsymbol{x}_i)}\\
D_{\text{KL}}(\boldsymbol{p}_2 || \boldsymbol{p}_1) = \sum\nolimits_{i=1}^{N}\sum\nolimits_{m=1}^{M}p_2^m(\boldsymbol{x}_i)\log\frac{p_2^m(\boldsymbol{x}_i)}{p_1^m(\boldsymbol{x}_i)}
\end{aligned}
\vspace{-5pt}
\end{equation}

%with $\boldsymbol{p}_1 = [\boldsymbol{p}_1^1, \boldsymbol{p}_1^2,\ldots,\boldsymbol{p}_1^m], \boldsymbol{p}_2 = [\boldsymbol{p}_2^1, \boldsymbol{p}_2^2,\ldots,\boldsymbol{p}_2^m]$

\noindent\textbf{Small-loss Selection} Before introducing the details, we first clarify the connection between small losses and clean instances. Intuitively, small-loss examples are likely to be the ones that are correctly labelled \cite{han2018co,ren2018learning}. Thus, if we train our classifier only using small-loss instances in each mini-batch data, it would be resistant to noisy labels.

\begin{algorithm}[!t]
\caption{JoCoR}
\label{alg:JoCoR}
\begin{algorithmic}[1]
\REQUIRE Network $f$ with $\boldsymbol{\Theta} = \{\boldsymbol{\Theta}_1, \boldsymbol{\Theta}_2\}$, learning rate $\eta$, fixed $\tau$, epoch $T_k$ and $T_{\text{max}}$, iteration $I_{\text{max}}$;
\FOR{$t$ = 1,2,\ldots,$T_{\text{max}}$}
    \STATE \textbf{Shuffle} training set $D$; %noisy dataset
    \FOR{$n = 1,\ldots,I_{\text{max}}$}
        \STATE \textbf{Fetch} mini-batch $D_n$ from $D$;
        \STATE $\boldsymbol{p}_1$ = $f(\boldsymbol{x},\boldsymbol{\Theta}_1)$, $\forall \boldsymbol{x} \in D_n$;
        \STATE $\boldsymbol{p}_2$ = $f(\boldsymbol{x},\boldsymbol{\Theta}_2)$, $\forall \boldsymbol{x} \in D_n$;

        %\STATE \textbf{Obtain} the predictions $p_1$ and $p_2$ for $\Tilde{D}$;
        \STATE \textbf{Calculate} the joint loss $\ell$ by \eqref{eq:joint_loss} using $\boldsymbol{p}_1$ and $\boldsymbol{p}_2$;
        \STATE \textbf{Obtain} small-loss sets $\Tilde{D}_n$ by \eqref{eq:small_loss} from $D_n$;
        \STATE \textbf{Obtain} $L$ by \eqref{eq:final_loss} on $\Tilde{D}_n$;

        \STATE \textbf{Update} $\boldsymbol{\Theta} = \boldsymbol{\Theta} - \eta \nabla L$;
    \ENDFOR
    \STATE \textbf{Update} $R(t) = 1 - \min \left\{ \frac{t}{T_k} \tau, \tau \right\}$
\ENDFOR
\ENSURE $\boldsymbol{\Theta}_1$ and $\boldsymbol{\Theta}_2$
\end{algorithmic}
%\vspace{-10pt}
\end{algorithm}

To handle noisy labels, we apply the ``small-loss" criterion to select ``clean" instances (step 8 in Algorithm \ref{alg:JoCoR}). Following the setting of Co-teaching+, we update $R(t)$ (step 12), which controls how many small-loss data should be selected in each training epoch. At the beginning of training, we keep more small-loss data (with a large $R(t)$) in each mini-natch since deep networks would fit clean data first \cite{arpit2017closer,zhang2016understanding}. With the increase of epochs, we reduce $R(t)$ gradually until reaching $1-\tau$, keeping fewer examples in each mini-batch. Such operation will prevent deep networks from over-fitting noisy data \cite{han2018co}.

In our algorithm, we use the joint loss \eqref{eq:joint_loss} to select small-loss examples. Intuitively, an instance with small joint loss means that both two networks could be easy to reach a consensus and make correct predictions on it. As two networks have different learning abilities based on different initial conditions, the selected small-loss instances are more likely to be with clean labels than those chosen by a single model. Specifically, we conduct small-loss selection as follows:

\begin{equation}
\begin{aligned}
\label{eq:small_loss}
\Tilde{D}_n = {\arg\min}_{D_n^\prime:\mid D_n^\prime \mid \ge R(t) \mid D_n \mid}\ell\left(D_n^\prime\right)
\end{aligned}
\vspace{5pt}
\end{equation}

After obtaining the small-loss instances, we calculate the average loss on these examples for further backpropagation:

\begin{equation}
\begin{aligned}
\label{eq:final_loss}
L = \frac{1}{|\Tilde{D}|}\sum\nolimits_{\boldsymbol{x} \in \Tilde{D}}\ell(\boldsymbol{x})
\end{aligned}
\vspace{5pt}
\end{equation}

\noindent\textbf{Relations to other approaches}. We compare JoCoR with other related approaches in Table \ref{tab:feature_compare}. Specifically, Decoupling applies the ``disagreement" strategy to select instances while Co-teaching use small-loss criterion. Besides, Co-teaching updates parameters of networks by the ``cross-update" strategy to reduce the accumulated error flow. Combining the ``disagreement" strategy and the ``cross-update" strategy, Co-teaching+ achieves excellent performance. As for our JoCoR, we also select small-loss examples but update the networks by Joint Training. Furthermore, we use the Co-Regularization to maximize agreement between the two networks. Note that Co-Regularization in our proposed method and ``disagreement” strategy in Decoupling are both essentially to reduce the divergence between the two classifiers. The difference between them lies in that the former uses an explicit regularization methods with all training examples while the latter employs hard sampling that reduces the effective number of training examples. It is especially important in the case of small-loss selection, because the selection would further decrease the effective number of training examples.

\begin{table}[!t]
\centering
\scriptsize
\topcaption{Comparison of state-of-the-art and related techniques with our JoCoR approach. In the first column, ``small loss": regarding small-loss samples as ``clean" samples, which is based on the memorization effects of deep neural networks; ``double classifiers": training two classifiers simultaneously; ``cross update": updating parameters in a cross manner instead of a parallel manner; ``joint training": training two classifiers with a joint loss; ``disagreement": updating two classifiers on disagreed examples during the entire training epochs; ``agreement": maximizing the agreement of two classifiers by regularization during the whole training epochs.}\label{tab:feature_compare}
\begin{tabular}{c|c|c|c|c}
\hline
\hline
 & Decoupling & Co-teaching & Co-teaching+ & JoCoR\\
\hline
small loss & \ding{55} & \ding{51} & \ding{51} & \ding{51} \\
\hline
cross update & \ding{55} & \ding{51} & \ding{51} & \ding{55} \\
\hline
joint training & \ding{55} & \ding{55} & \ding{55} & \ding{51} \\
\hline
disagreement & \ding{51} & \ding{55} & \ding{51} & \ding{55} \\
\hline
agreement & \ding{55} & \ding{55} & \ding{55} & \ding{51} \\
\hline
\hline
\end{tabular}
\vspace{-10pt}
\end{table}

\begin{figure*}[ht]
    \centering
    \subfigure{
        \centering
        \includegraphics[scale=0.36]{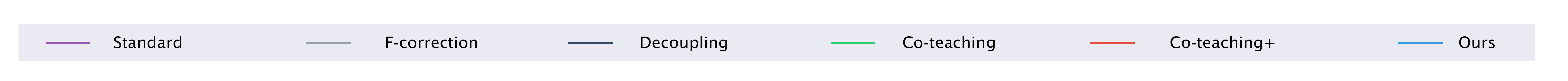}
    }
    
    \vspace{-16pt}

    \subfigure{
        \centering
        \includegraphics[width=4cm,height=3.5cm]{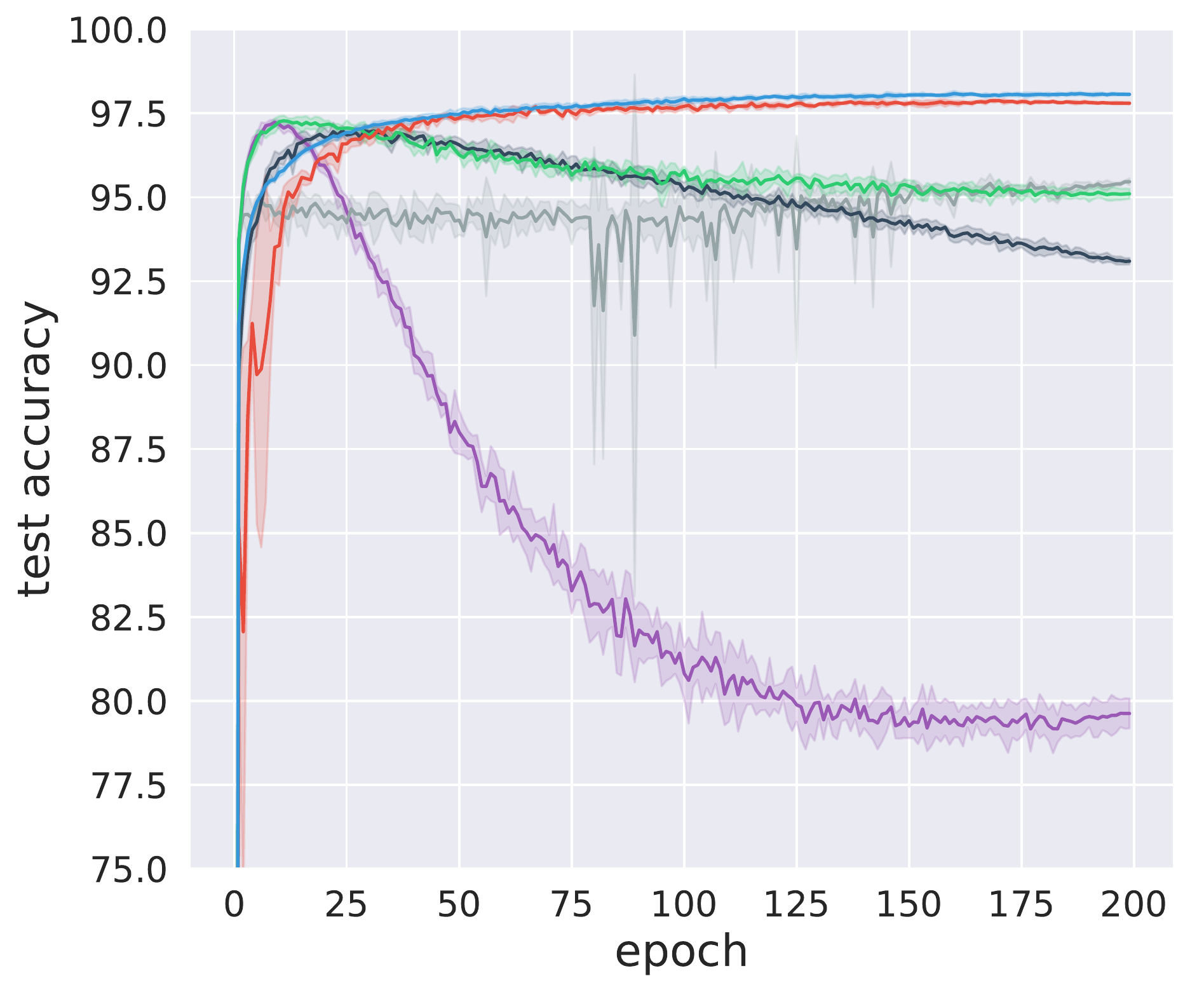}
    }
    \subfigure{
        \centering
        \includegraphics[width=4cm,height=3.5cm]{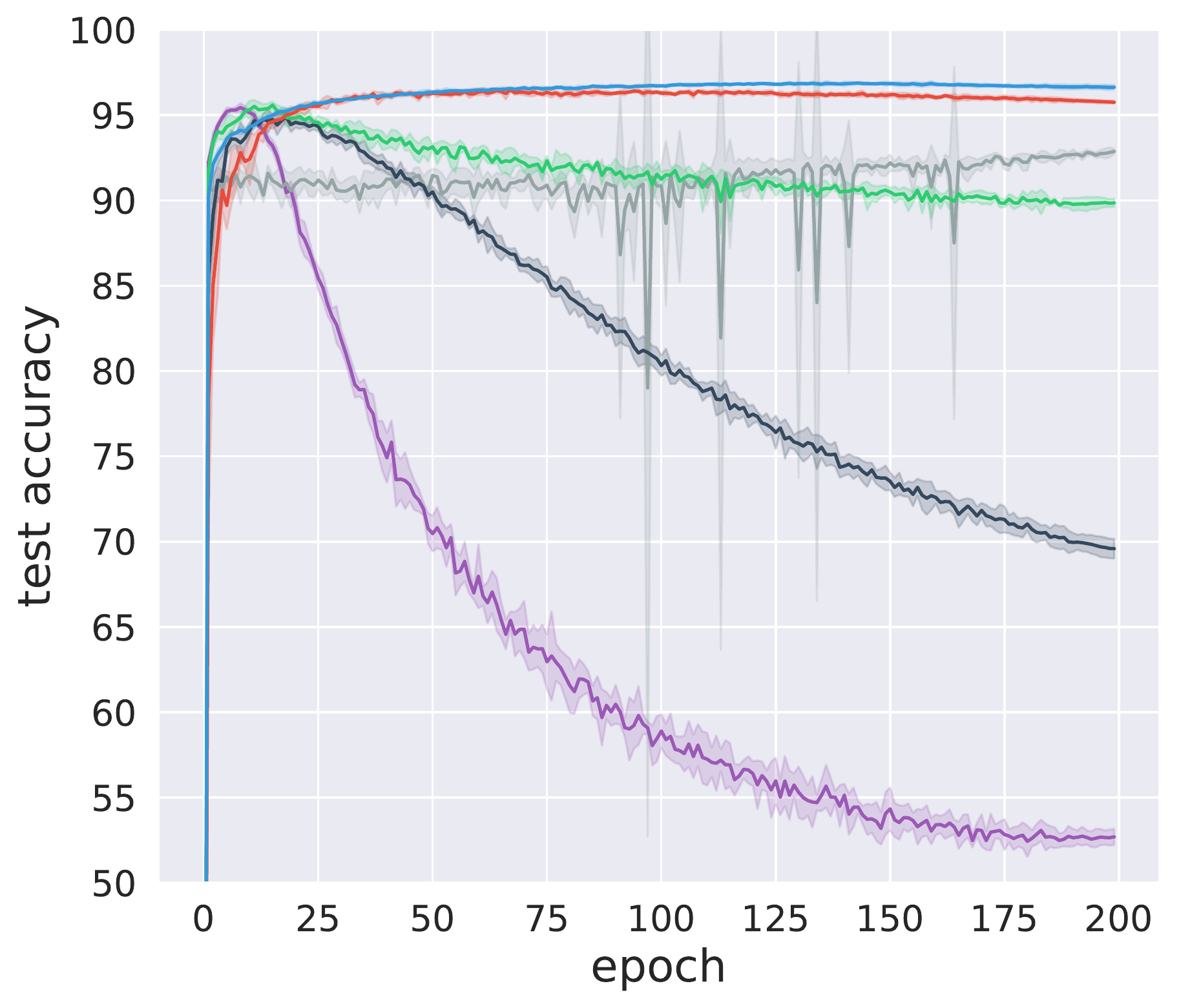}
    }
    \subfigure{
        \centering
        \includegraphics[width=4cm,height=3.5cm]{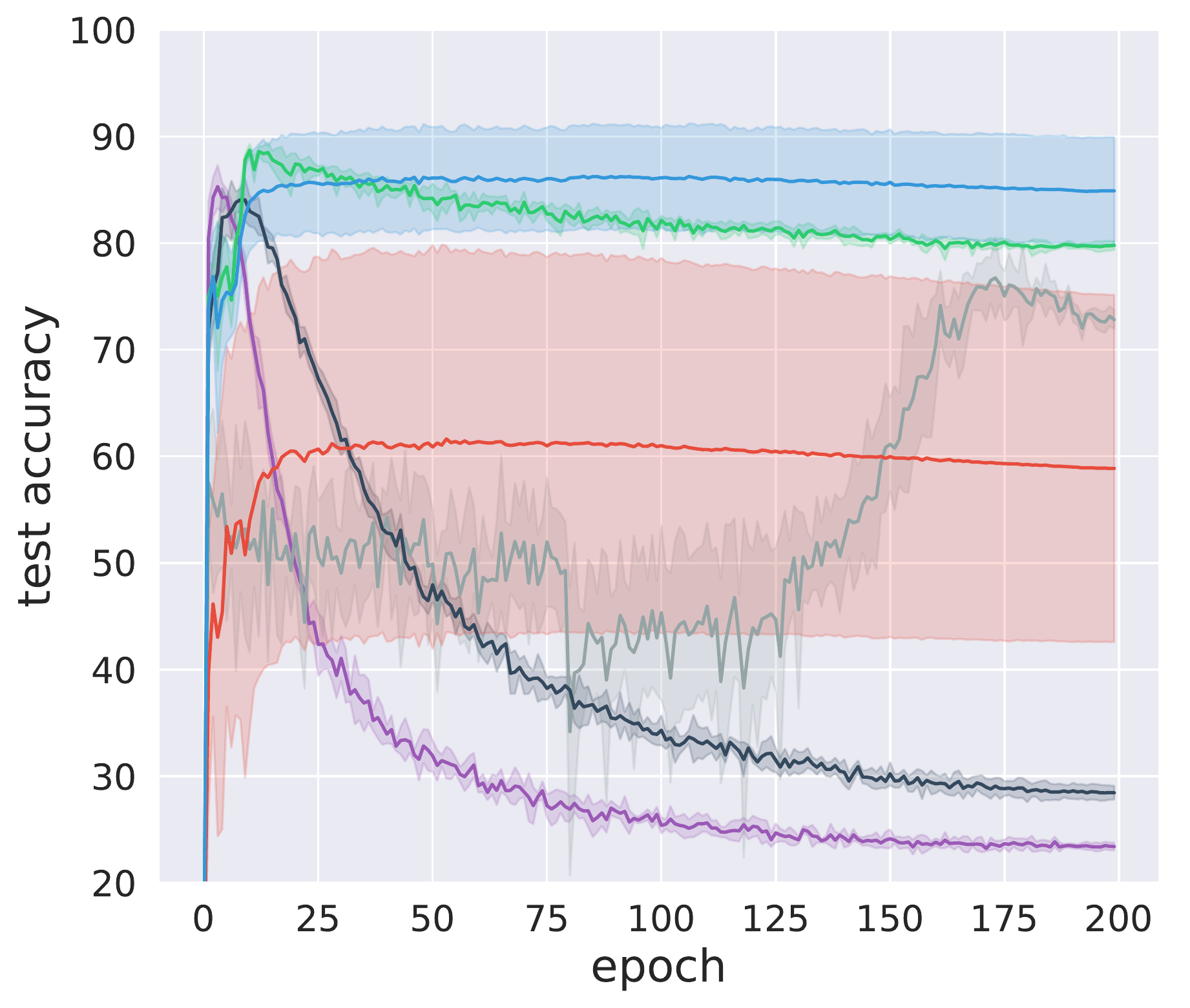}
    }
    \subfigure{
        \centering
        \includegraphics[width=4cm,height=3.5cm]{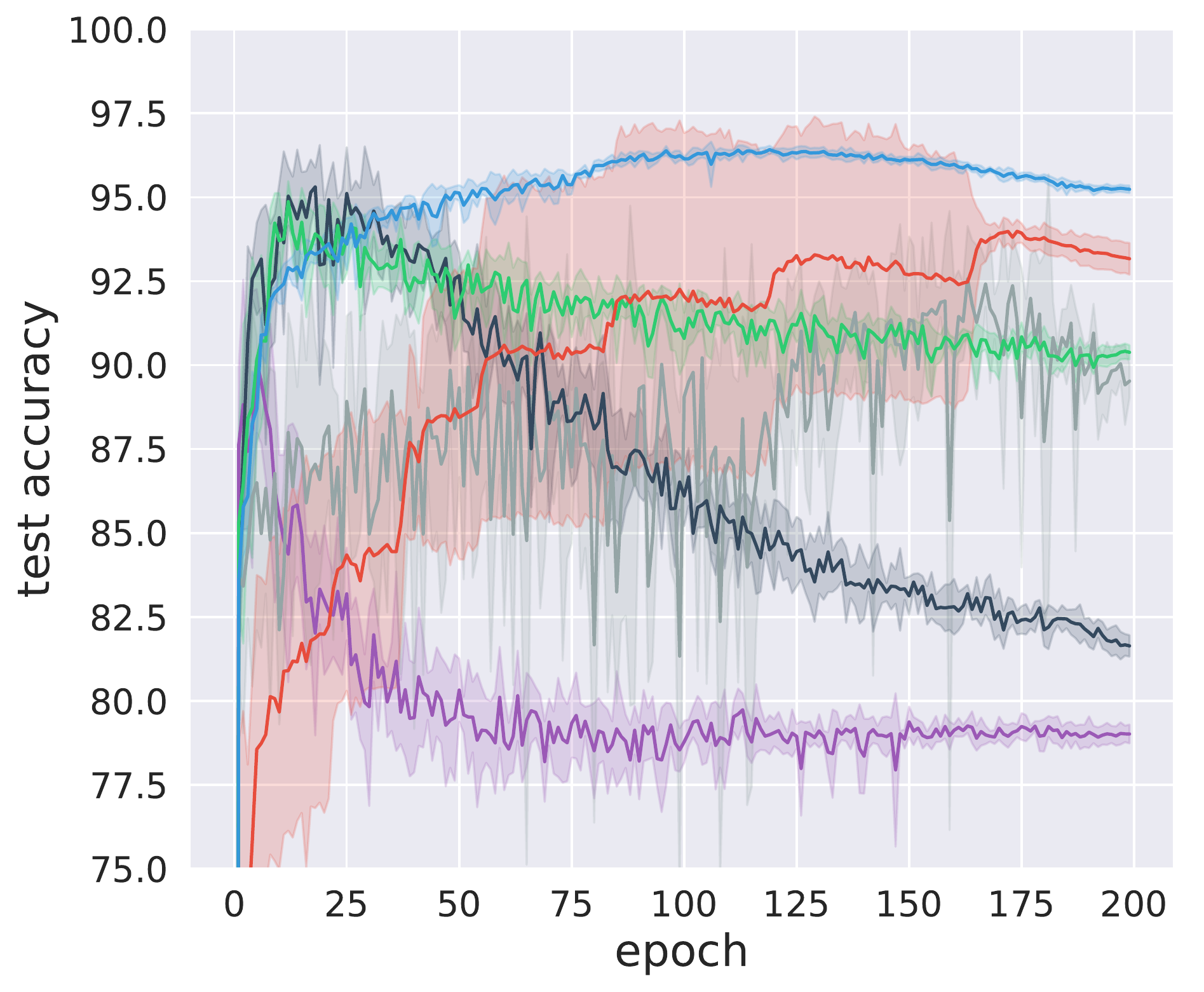}
    }
    
    \vspace{-10pt}
    \setcounter{subfigure}{0}
    \subfigure[Symmetry-20\%]  {
        \centering
        \includegraphics[width=4cm,height=3cm]{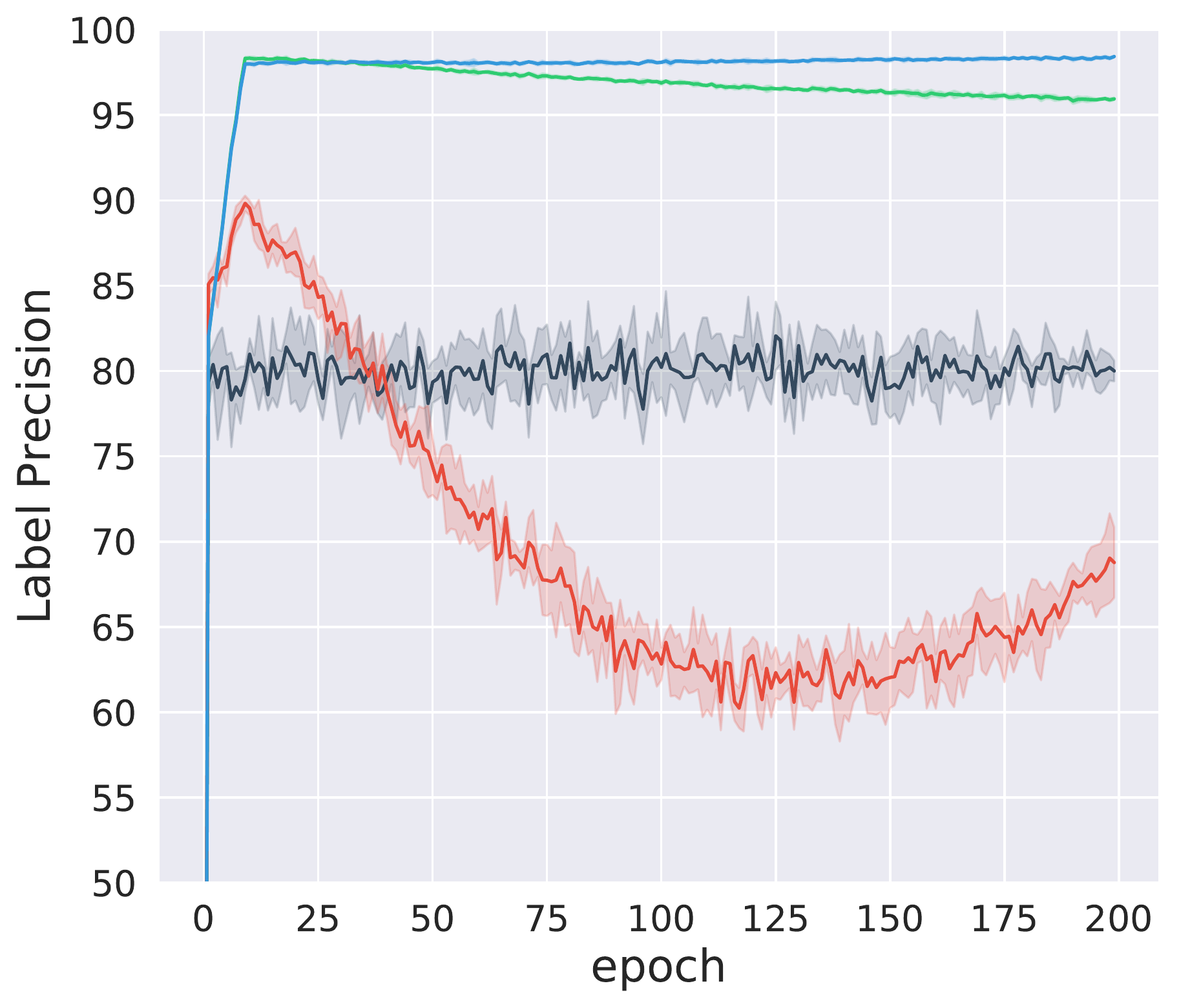}
    }
    \subfigure[Symmetry-50\%]{
        \centering
        \includegraphics[width=4cm,height=3cm]{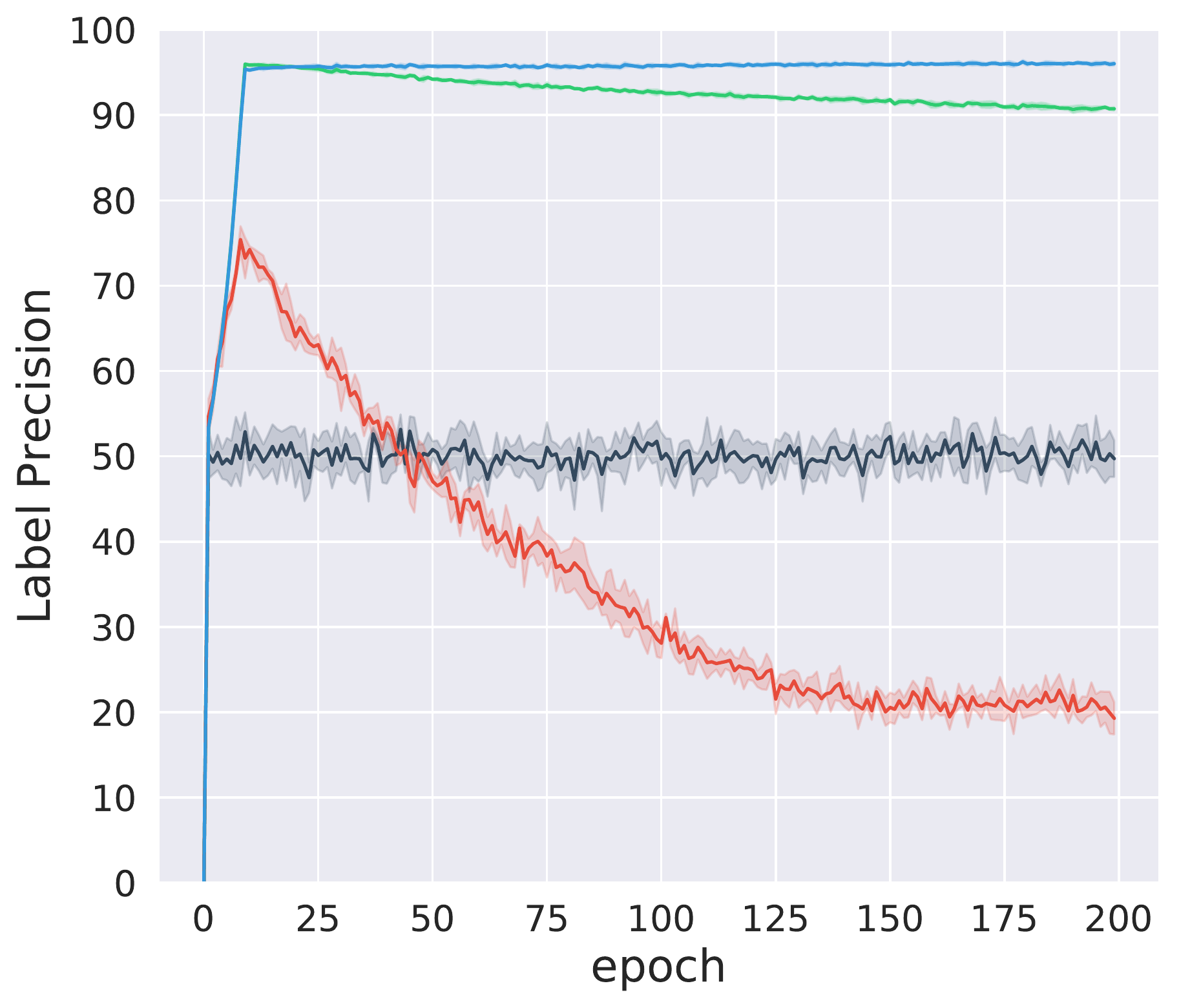}
    }
    \subfigure[Symmetry-80\%]{
        \centering
        \includegraphics[width=4cm,height=3cm]{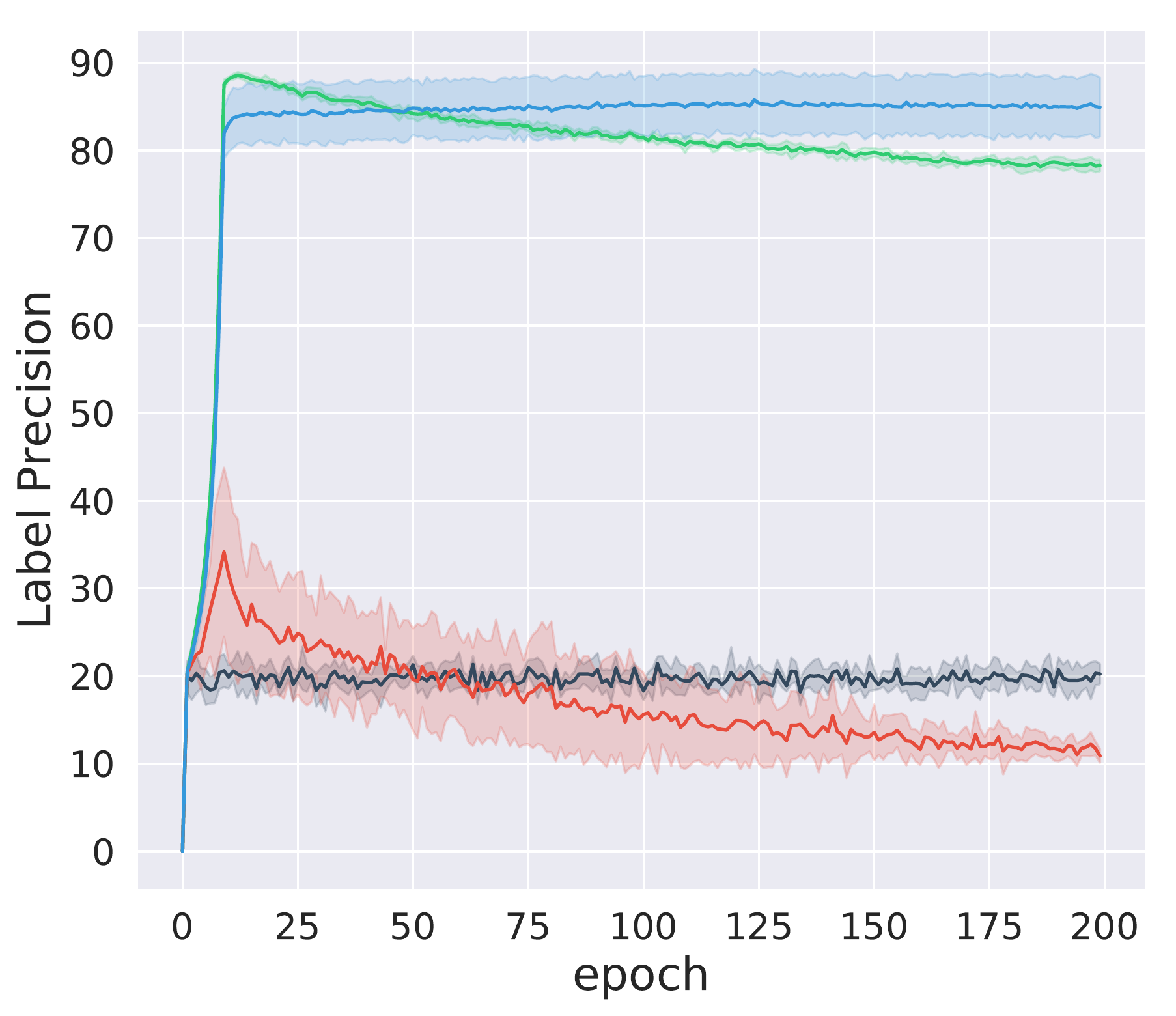}
    }
    \subfigure[Asymmetry-40\%]{
        \centering
        \includegraphics[width=4cm,height=3cm]{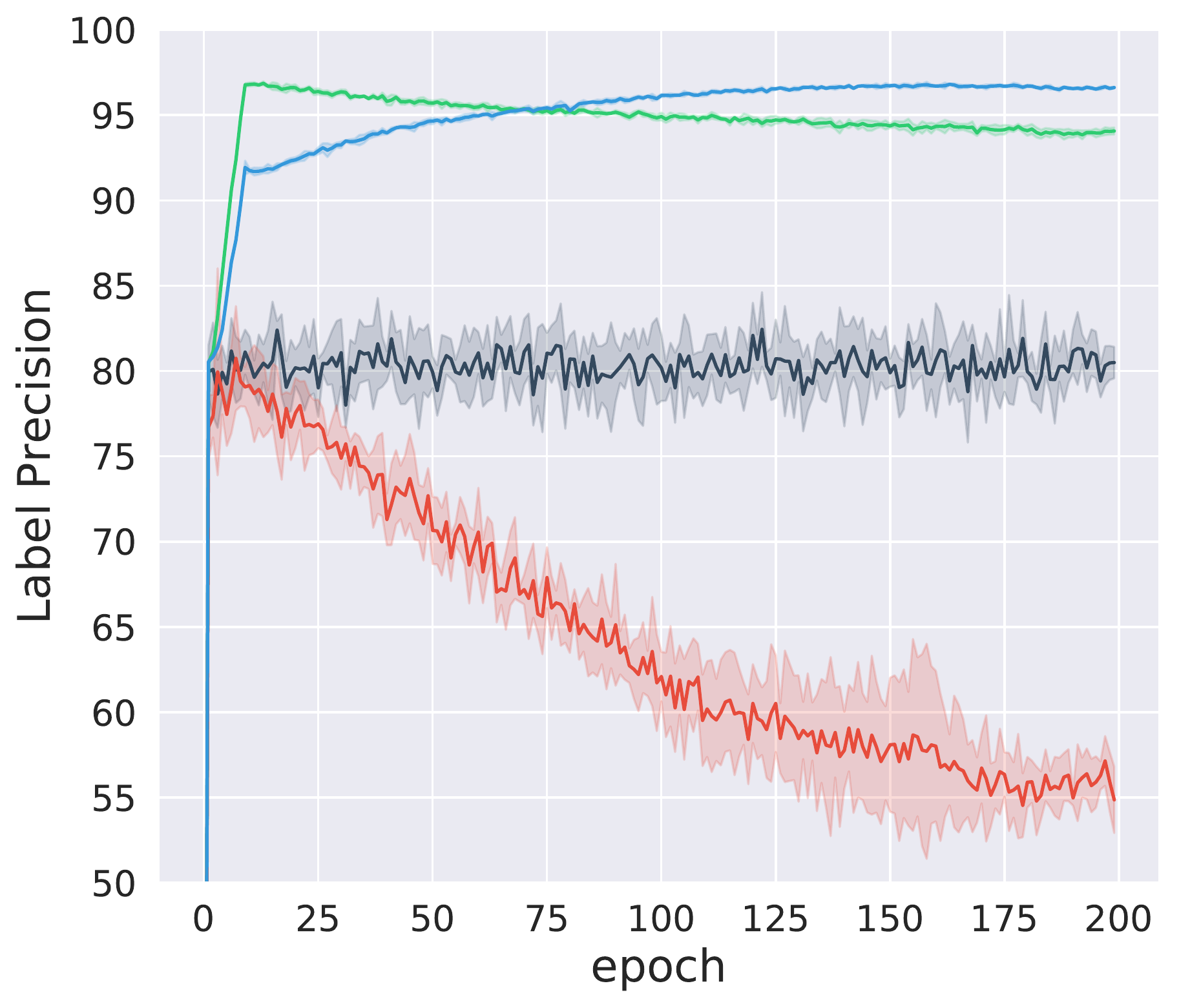}
    }
    \caption{Results on MNIST dataset. Top: test accuracy(\%) vs. epochs; bottom: label precision(\%) vs. epochs.}
    \vspace{-5pt}
    \label{fig:mnist_performance}
\end{figure*}

\begin{table*}[ht]
\centering

\topcaption{Average test accuracy (\%) on \textsl{MNIST} over the last 10 epochs.}\label{tab:mnist_data}
\begin{tabular}{c|c|c|c|c|c|c}
\hline
\hline
Flipping-Rate& Standard & F-correction & Decoupling & Co-teaching & Co-teaching+ & JoCoR\\
\hline
Symmetry-20\% &  $79.56 \pm 0.44$ & $95.38 \pm 0.10$ & $93.16 \pm 0.11$ & $95.10 \pm 0.16$ & $97.81 \pm 0.03$ & $\textbf{98.06} \pm 0.04$ \\
\hline
Symmetry-50\% &  $52.66 \pm 0.43$ & $92.74 \pm 0.21$ & $69.79 \pm 0.52$ & $89.82 \pm 0.31$ & $95.80 \pm 0.09$ & $\textbf{96.64} \pm 0.12$ \\
\hline
Symmetry-80\% &  $23.43 \pm 0.31$ & $72.96 \pm 0.90$ & $28.51 \pm 0.65$ & $79.73 \pm 0.35$ & $58.92 \pm 14.73$ & $\textbf{84.89} \pm 4.55$ \\
\hline
Asymmetry-40\% & $79.00 \pm 0.28$ & $89.77 \pm 0.96$ & $81.84 \pm 0.38$ & $90.28 \pm 0.27$ & $93.28 \pm 0.43$ & $\textbf{95.24} \pm 0.10$ \\
\hline
\hline
\end{tabular}
\vspace*{-17pt}

\end{table*}

\begin{figure}[ht]
    \centering
    \subfigure{
        \centering
        \includegraphics[scale=0.32]{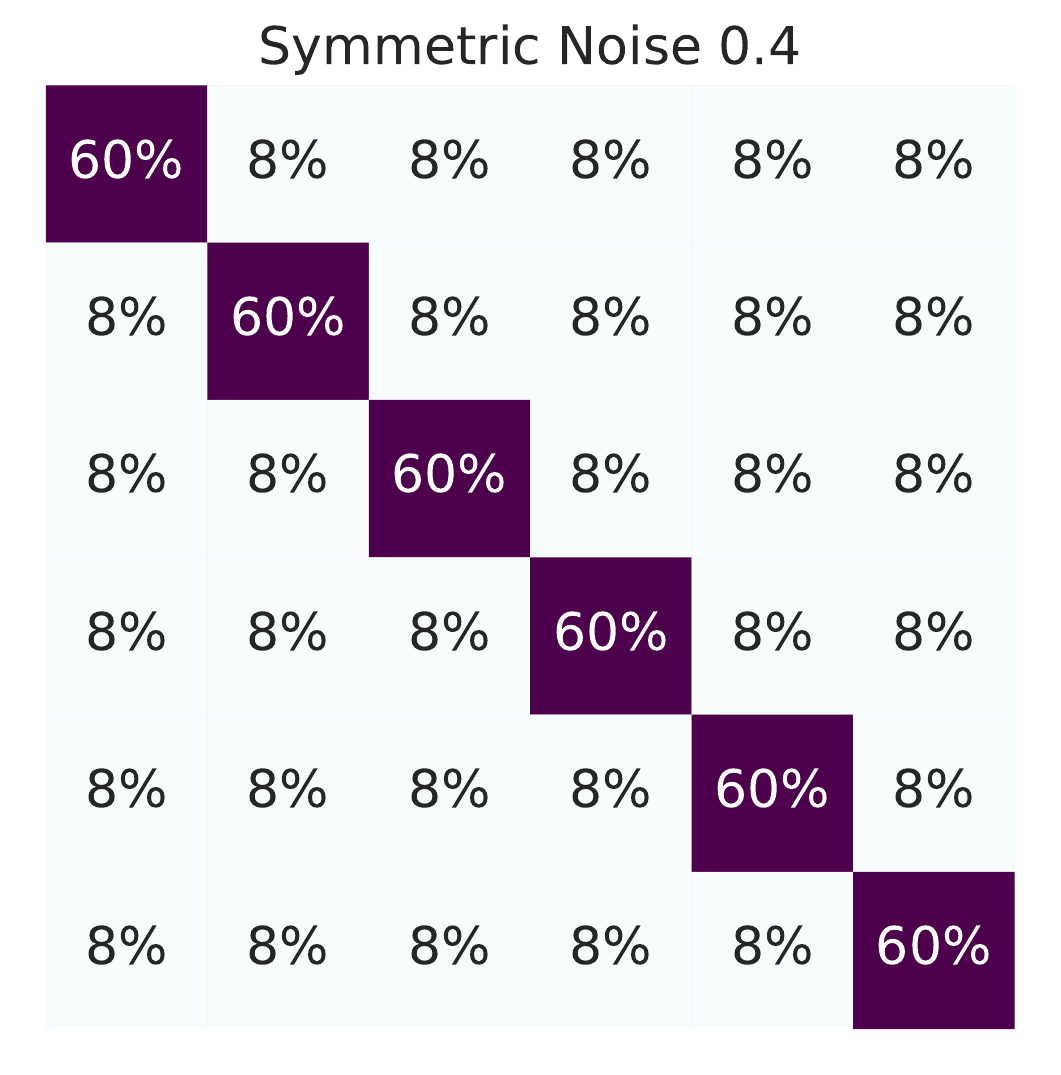}
    }
    \subfigure{
        \centering
        \includegraphics[scale=0.32]{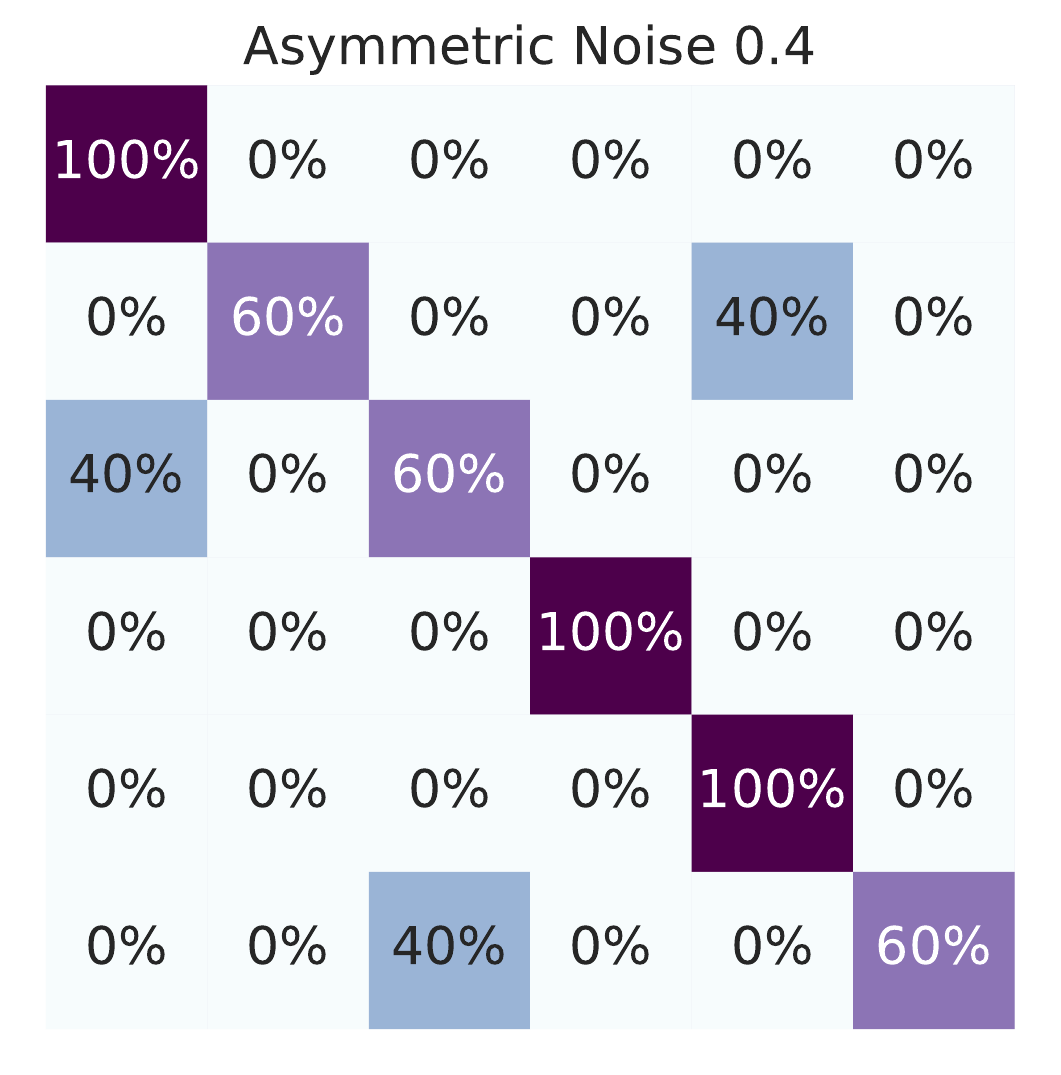}
    }
    \vspace{-2pt}
    \caption{Example of noise transition matrix $T$ (taking 6 classes and noise ratio 0.4 as an example)}
    \vspace{-18pt}
    \label{fig:matrix}
\end{figure}

\begin{figure*}[ht]
    \centering
    \subfigure{
    \begin{minipage}[t]{\textwidth}
        \centering
        \includegraphics[scale=0.36]{img/legend.pdf}
    \end{minipage}
    }
    
    \vspace{-16pt}

    \subfigure{
    \begin{minipage}[t]{0.2\textwidth}
        \centering
        \includegraphics[scale=0.2]{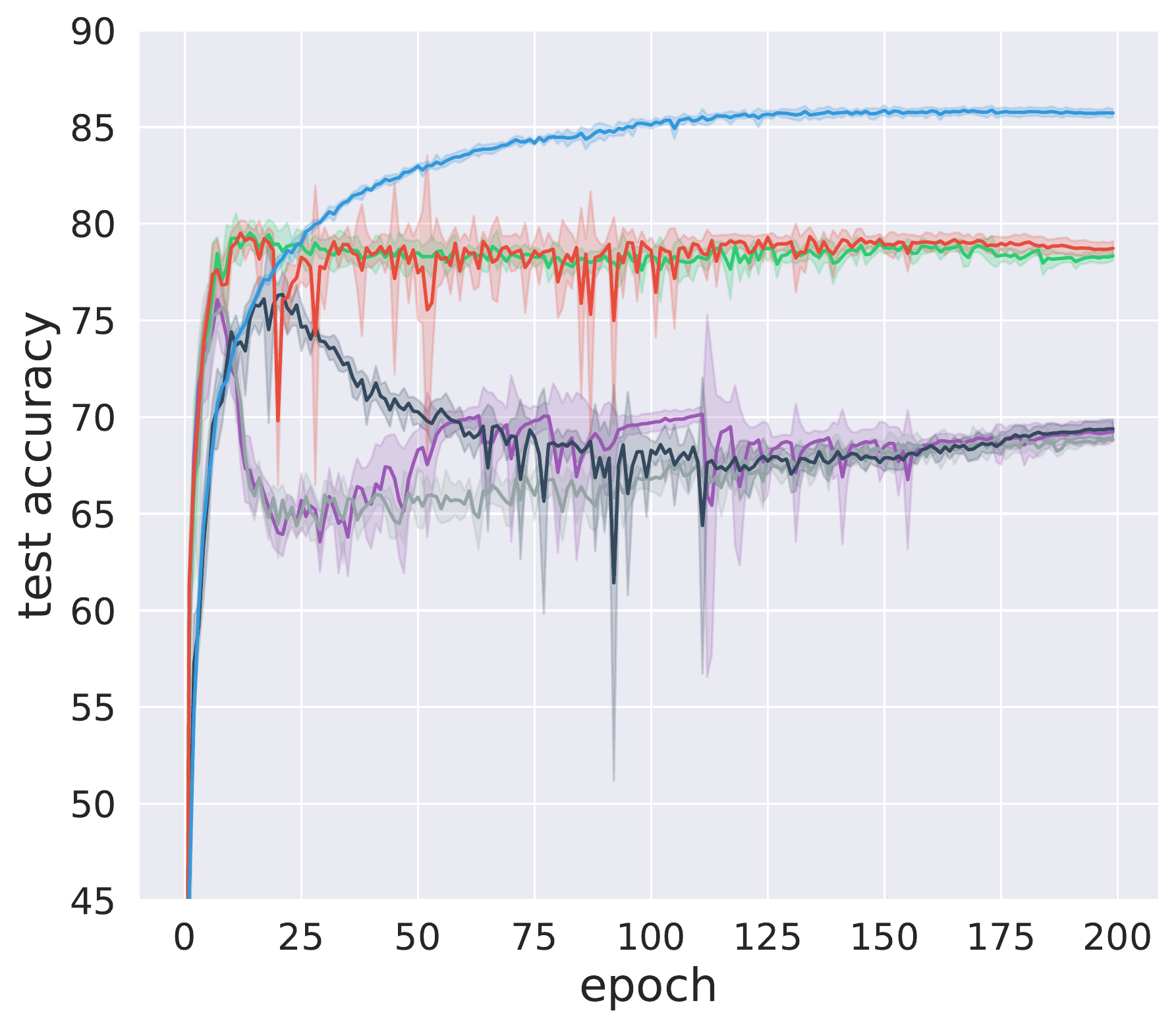}
    \end{minipage}
    }
    \subfigure{
    \begin{minipage}[t]{0.2\textwidth}
        \centering
        \includegraphics[scale=0.2]{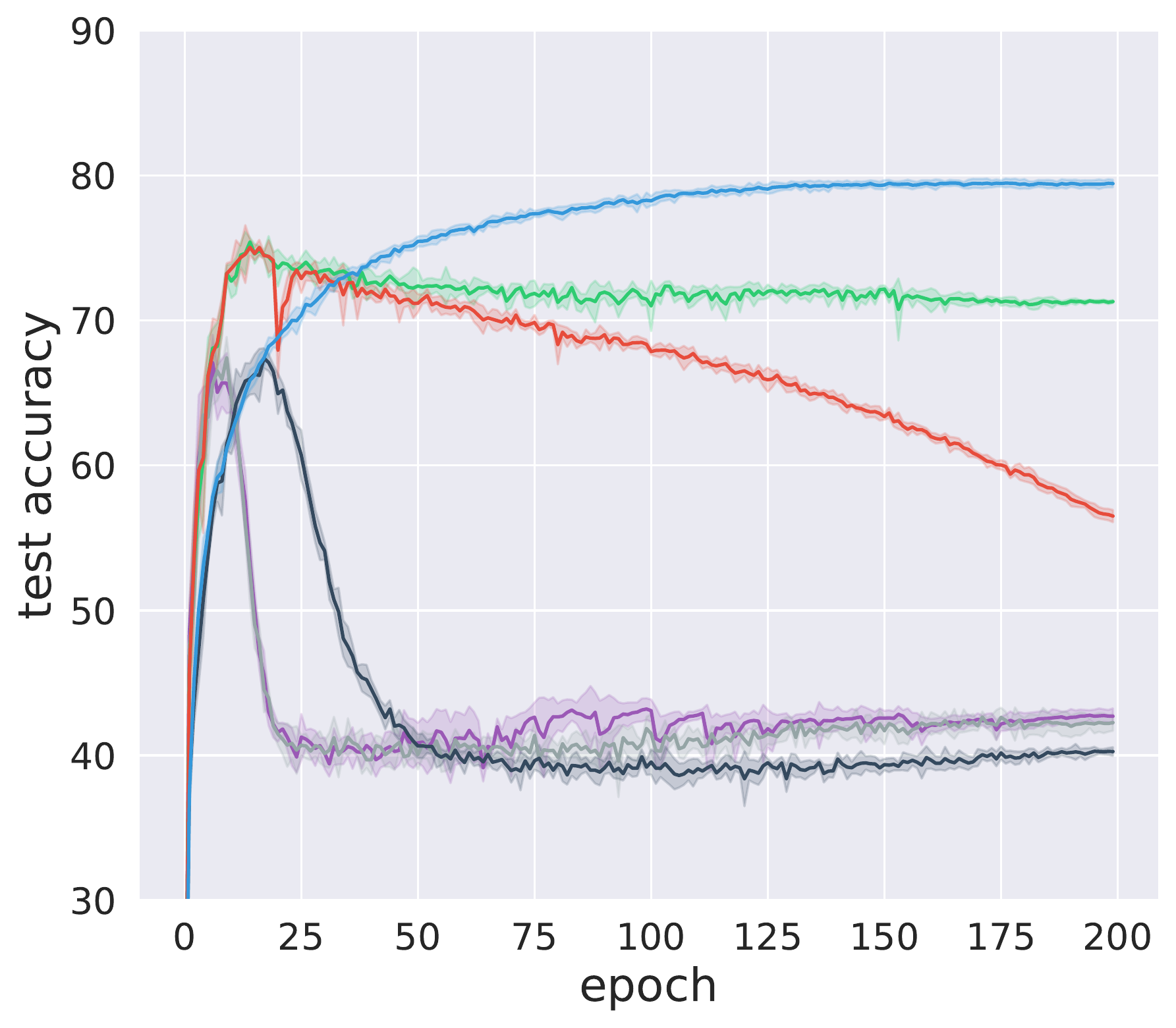}
    \end{minipage}
    }
    \subfigure{
    \begin{minipage}[t]{0.2\textwidth}
        \centering
        \includegraphics[scale=0.2]{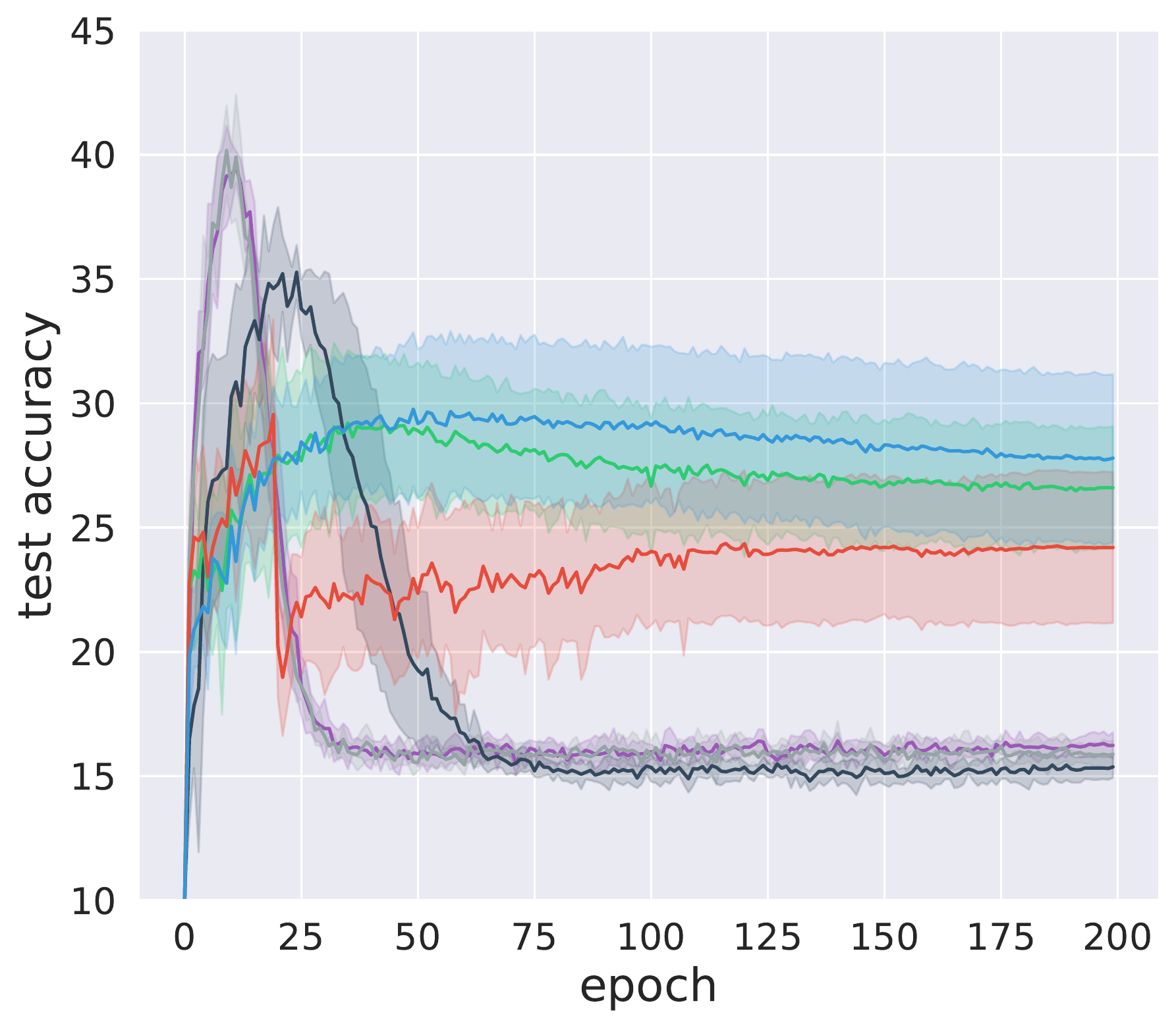}
    \end{minipage}
    }
    \subfigure{
    \begin{minipage}[t]{0.2\textwidth}
        \centering
        \includegraphics[scale=0.2]{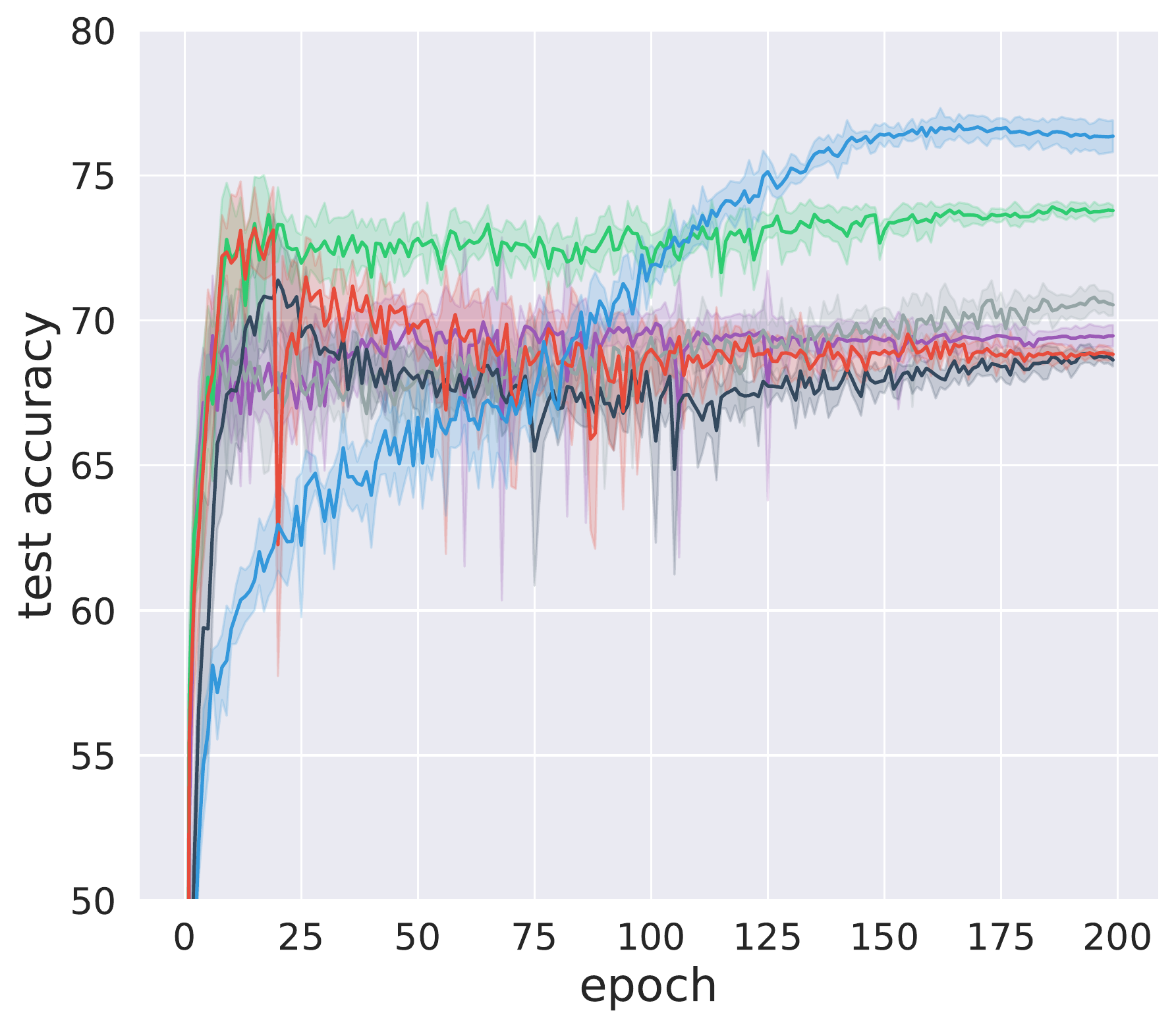}
    \end{minipage}
    }
    
    \vspace{-10pt}
    \setcounter{subfigure}{0}

    \subfigure[Symmetry-20\%]  {
    \begin{minipage}[t]{0.2\textwidth}
        \centering
        \includegraphics[scale=0.2]{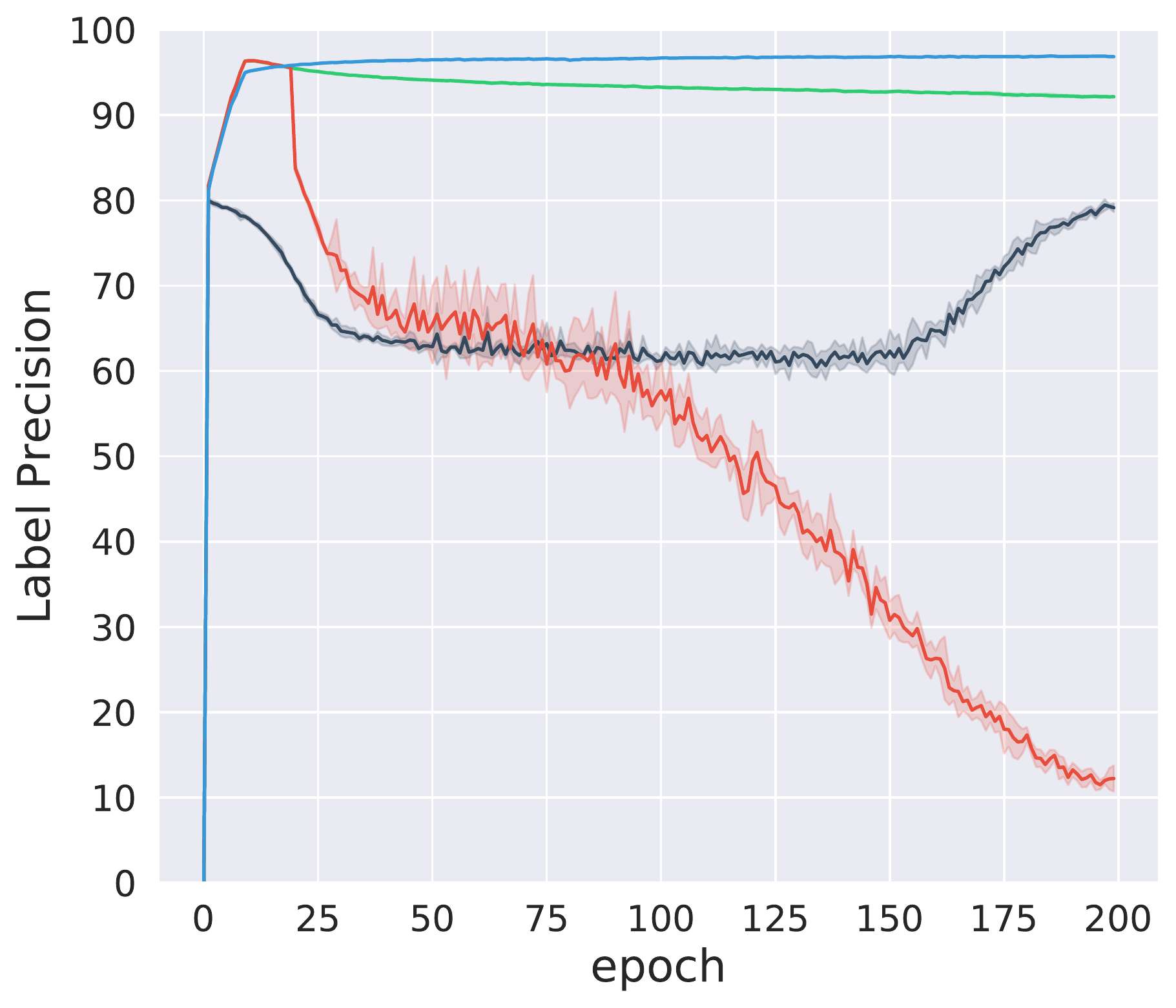}
    \end{minipage}
    }
    \subfigure[Symmetry-50\%]{
    \begin{minipage}[t]{0.2\textwidth}
        \centering
        \includegraphics[scale=0.2]{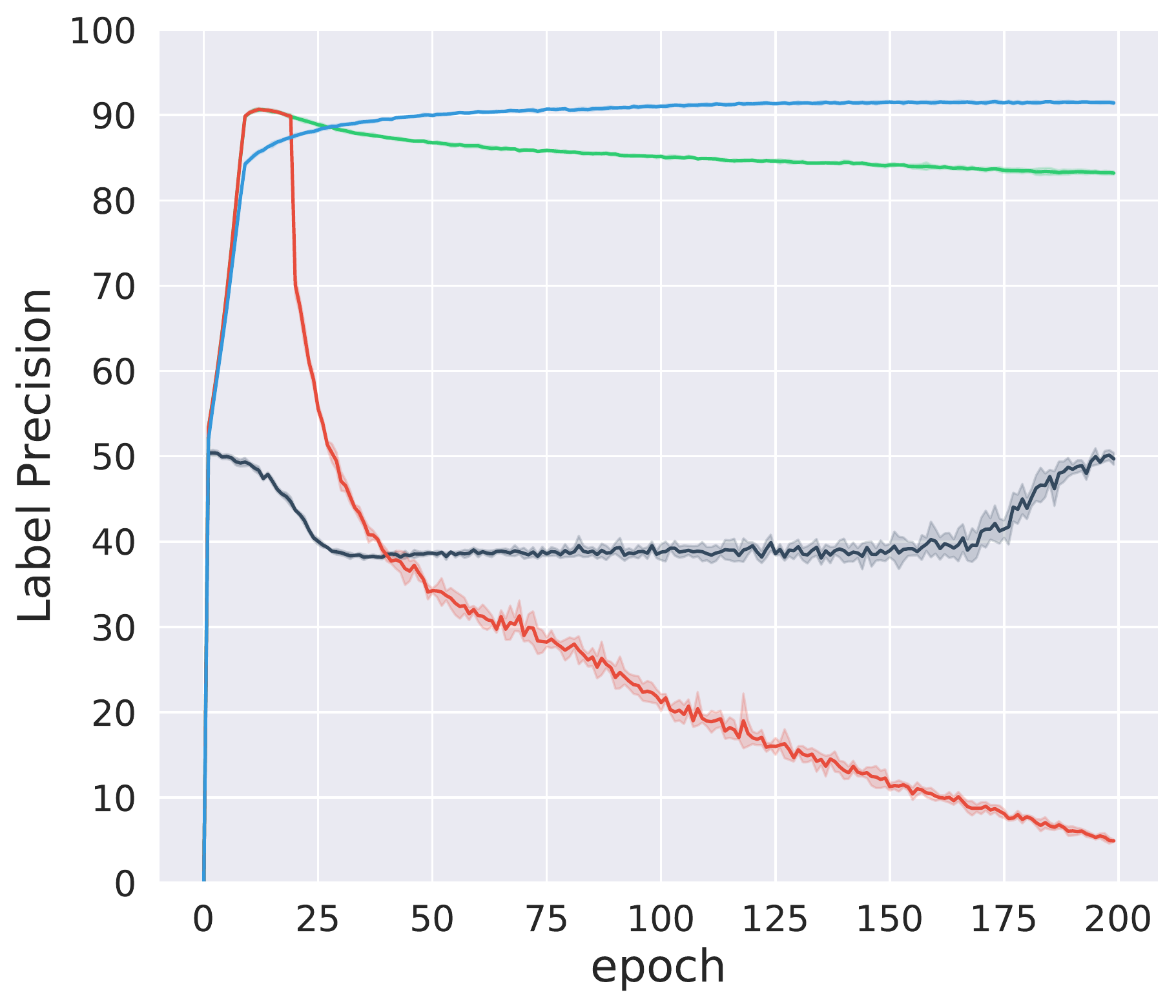}
    \end{minipage}
    }
    \subfigure[Symmetry-80\%]{
    \begin{minipage}[t]{0.2\textwidth}
        \centering
        \includegraphics[scale=0.2]{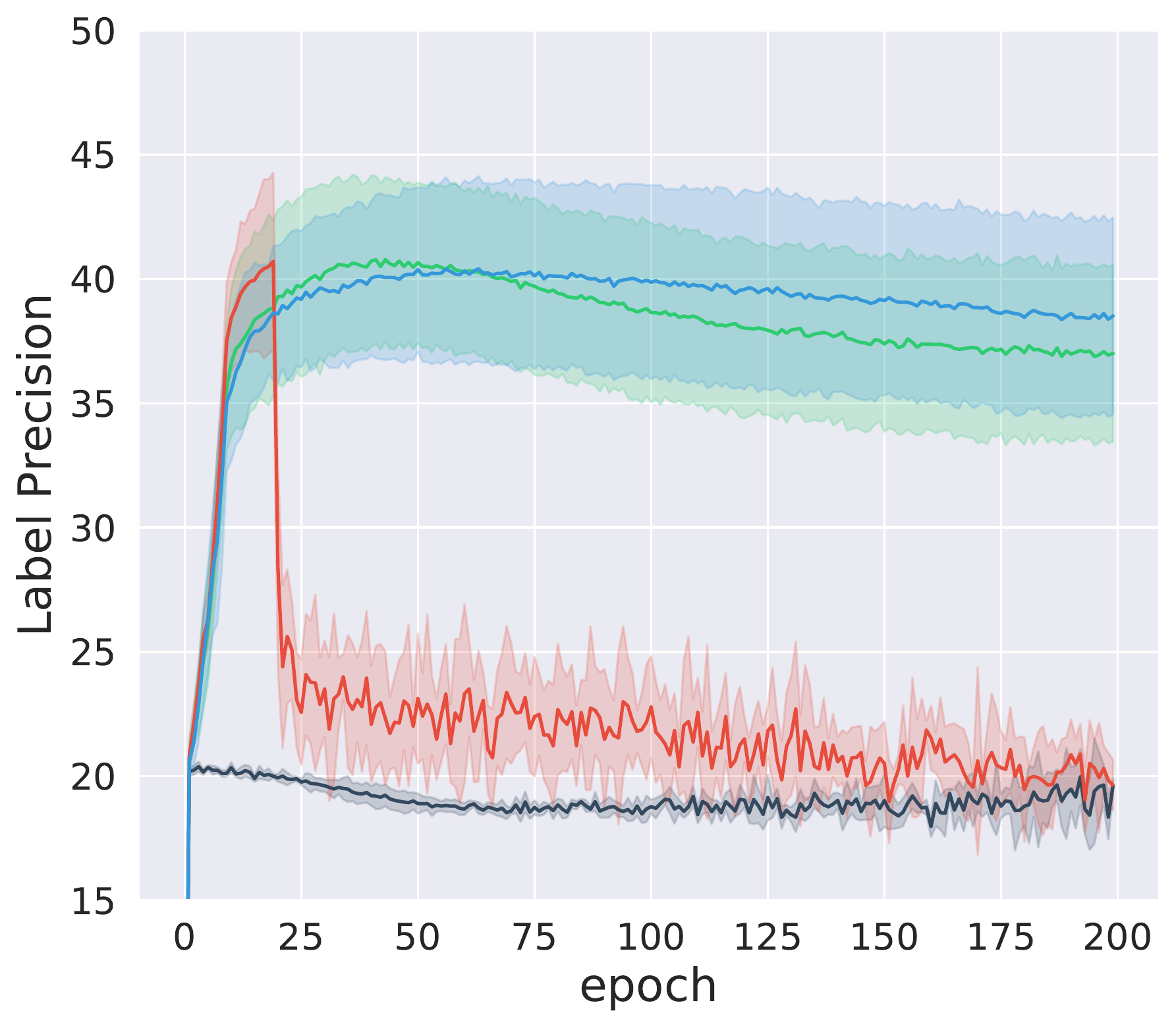}
    \end{minipage}
    }
    \subfigure[Asymmetry-40\%]{
    \begin{minipage}[t]{0.2\textwidth}
        \centering
        \includegraphics[scale=0.2]{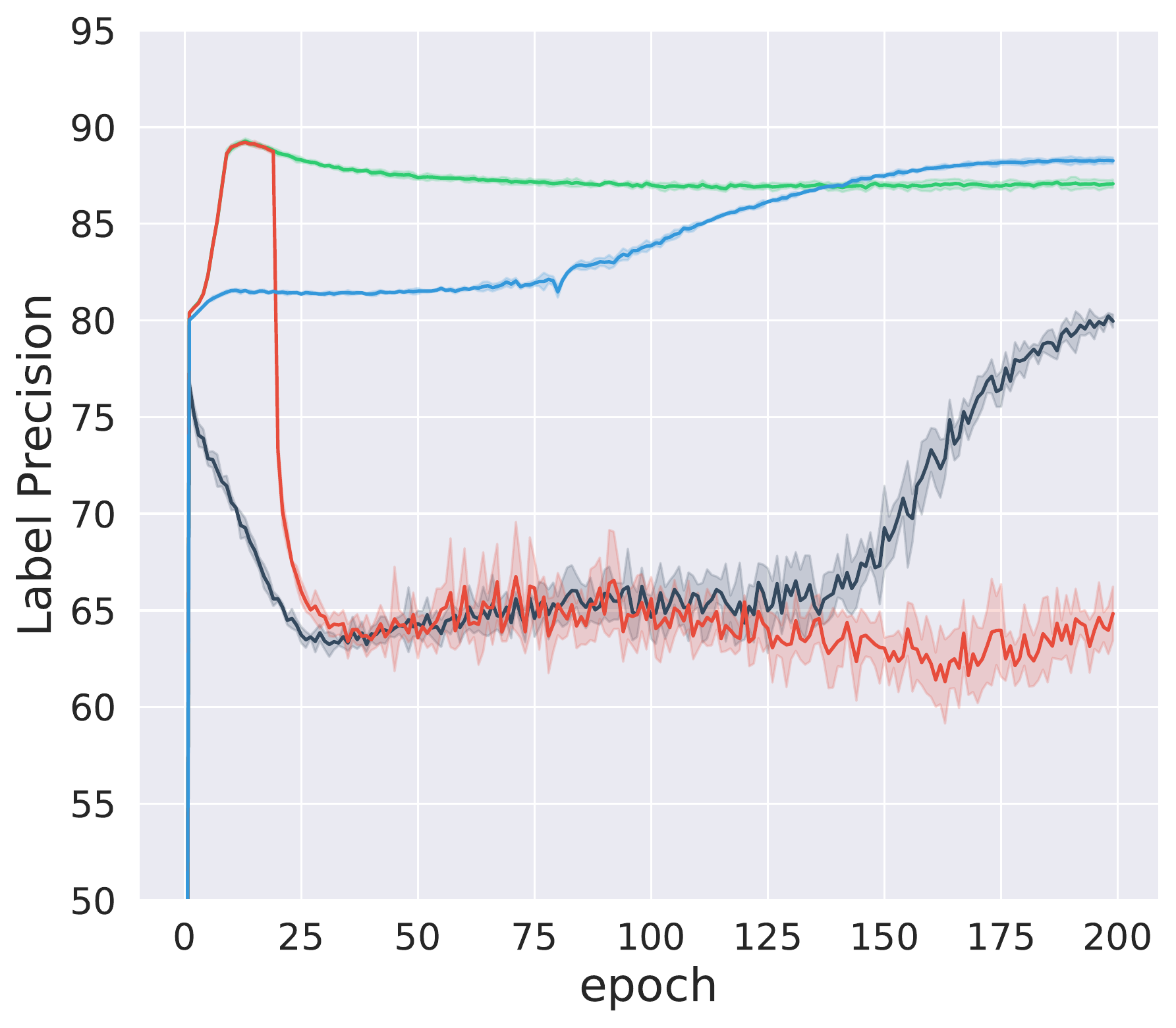}
    \end{minipage}
    }
    \caption{Results on CIFAR-10 dataset. Top: test accuracy(\%) vs. epochs; bottom: label precision(\%) vs. epochs.}
    \vspace{-5pt}
    \label{fig:cifar10_performance}
    
\end{figure*}

\begin{table*}[ht]
\centering
\topcaption{Average test accuracy (\%) on \textsl{CIFAR-10} over the last 10 epochs.}\label{tab:cifar10_data}
\begin{tabular}{c|c|c|c|c|c|c}
\hline
\hline
Flipping-Rate& Standard & F-correction & Decoupling & Co-teaching & Co-teaching+ & JoCoR\\
\hline
Symmetry-20\% &  $69.18 \pm 0.52$ & $68.74 \pm 0.20$ & $69.32 \pm 0.40$ & $78.23 \pm 0.27$ & $78.71 \pm 0.34$ & $\textbf{85.73} \pm 0.19$ \\
\hline
Symmetry-50\% &  $42.71 \pm 0.42$ & $42.19 \pm 0.60$ & $40.22 \pm 0.30$ & $71.30 \pm 0.13$ & $57.05 \pm 0.54$ & $\textbf{79.41} \pm 0.25$ \\
\hline
Symmetry-80\% &  $16.24 \pm 0.39$ & $15.88 \pm 0.42$ & $15.31 \pm 0.43$ & $26.58 \pm 2.22$ & $24.19 \pm 2.74$ & $\textbf{27.78} \pm 3.06$ \\
\hline
Asymmetry-40\% &  $69.43 \pm 0.33$ & $70.60\pm 0.40$ & $68.72 \pm 0.30$ & $73.78 \pm 0.22$ & $68.84 \pm 0.20$ & $\textbf{76.36} \pm 0.49$ \\
\hline
\hline
\end{tabular}
\vspace*{-15pt}
\end{table*}

\begin{figure*}[ht]
    \centering
    \subfigure{
    \begin{minipage}[t]{\textwidth}
        \centering
        \includegraphics[scale=0.36]{img/legend.pdf}
    \end{minipage}
    }
    
    \vspace{-16pt}

    \subfigure{
    \begin{minipage}[t]{0.2\textwidth}
        \centering
        \includegraphics[scale=0.2]{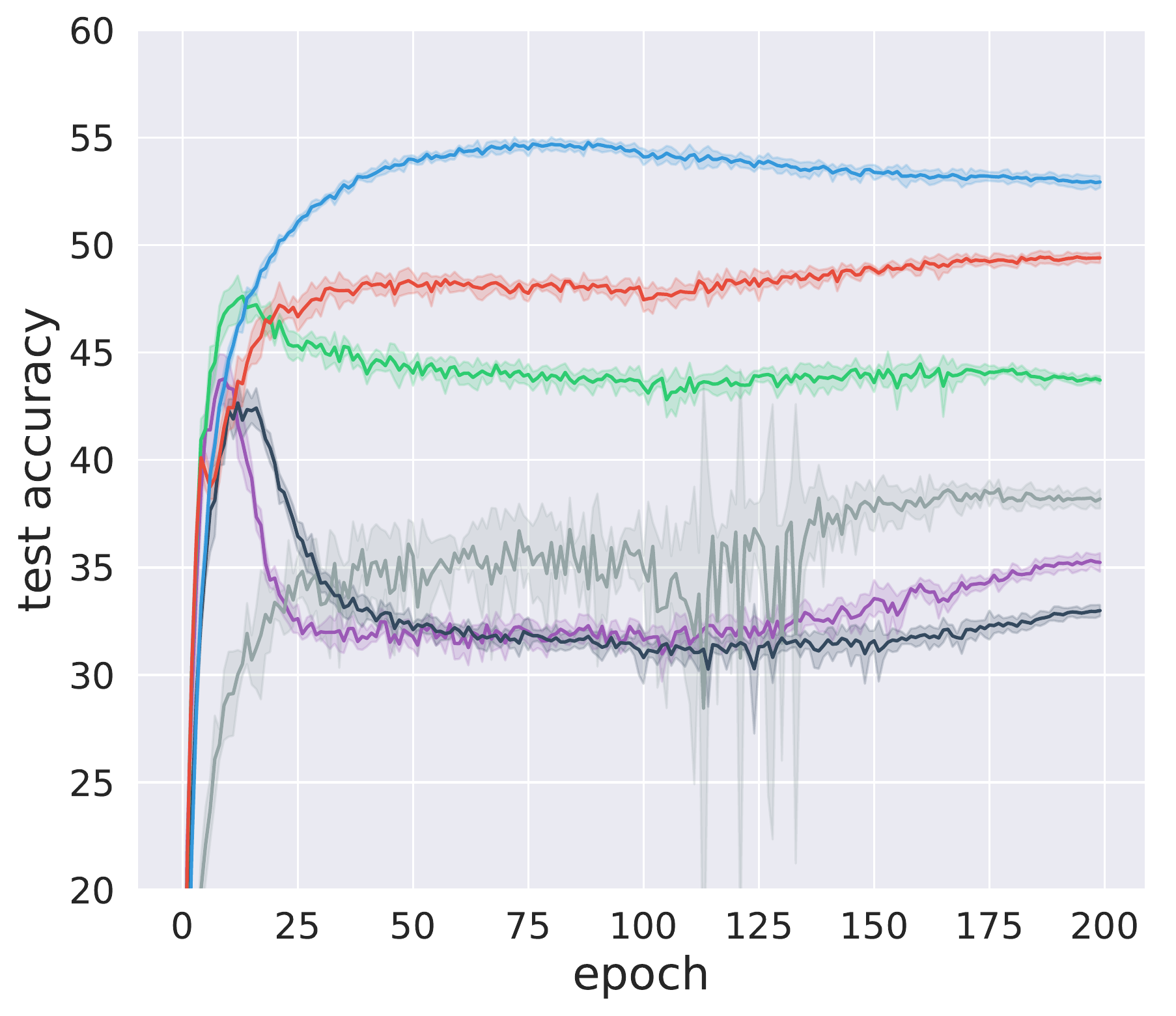}
    \end{minipage}
    }
    \subfigure{
    \begin{minipage}[t]{0.2\textwidth}
        \centering
        \includegraphics[scale=0.2]{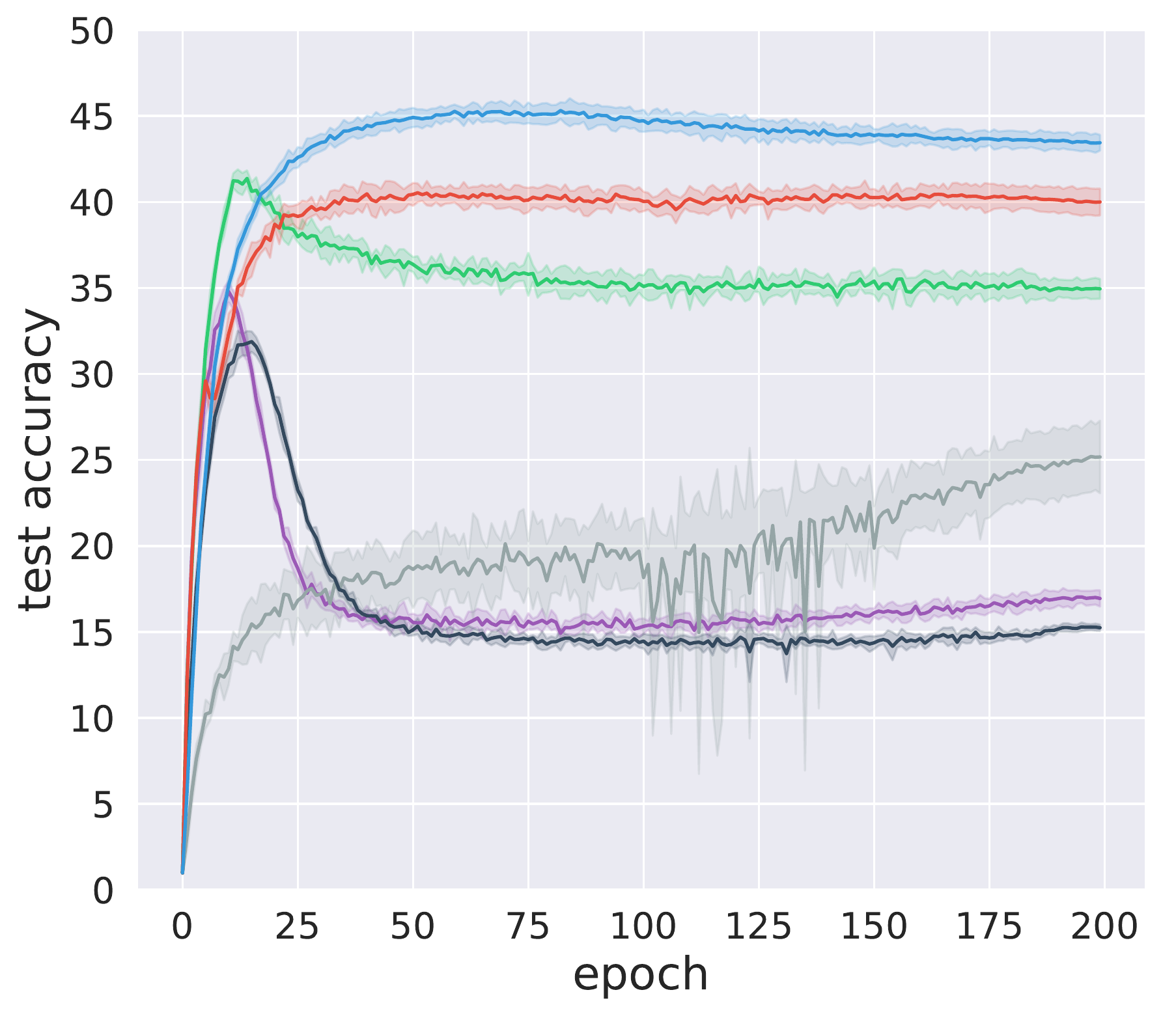}
    \end{minipage}
    }
    \subfigure{
    \begin{minipage}[t]{0.2\textwidth}
        \centering
        \includegraphics[scale=0.2]{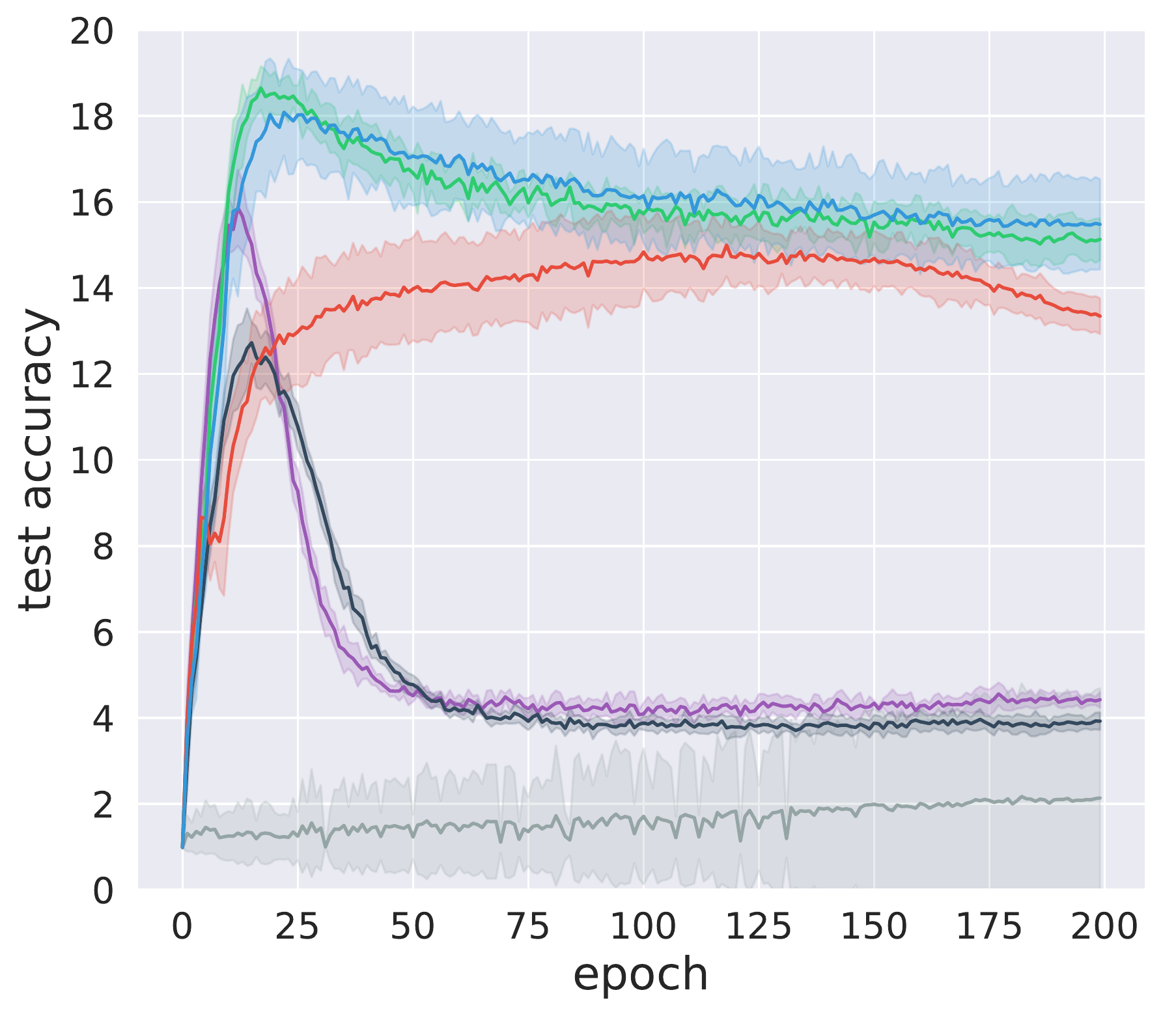}
    \end{minipage}
    }
    \subfigure{
    \begin{minipage}[t]{0.2\textwidth}
        \centering
        \includegraphics[scale=0.2]{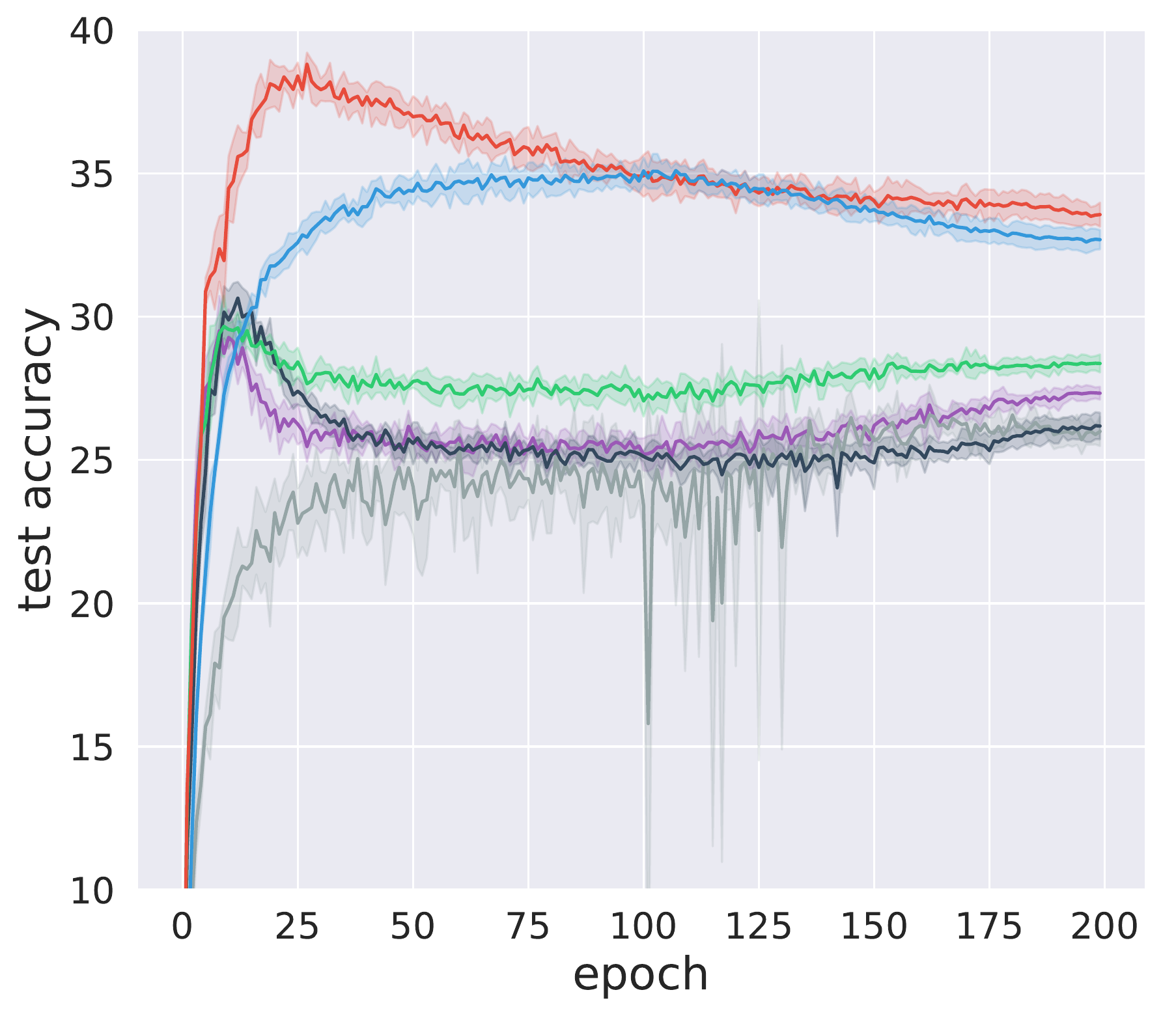}
    \end{minipage}
    }
    
    \vspace{-10pt}
    \setcounter{subfigure}{0}

    \subfigure[Symmetry-20\%]  {
    \begin{minipage}[t]{0.2\textwidth}
        \centering
        \includegraphics[scale=0.2]{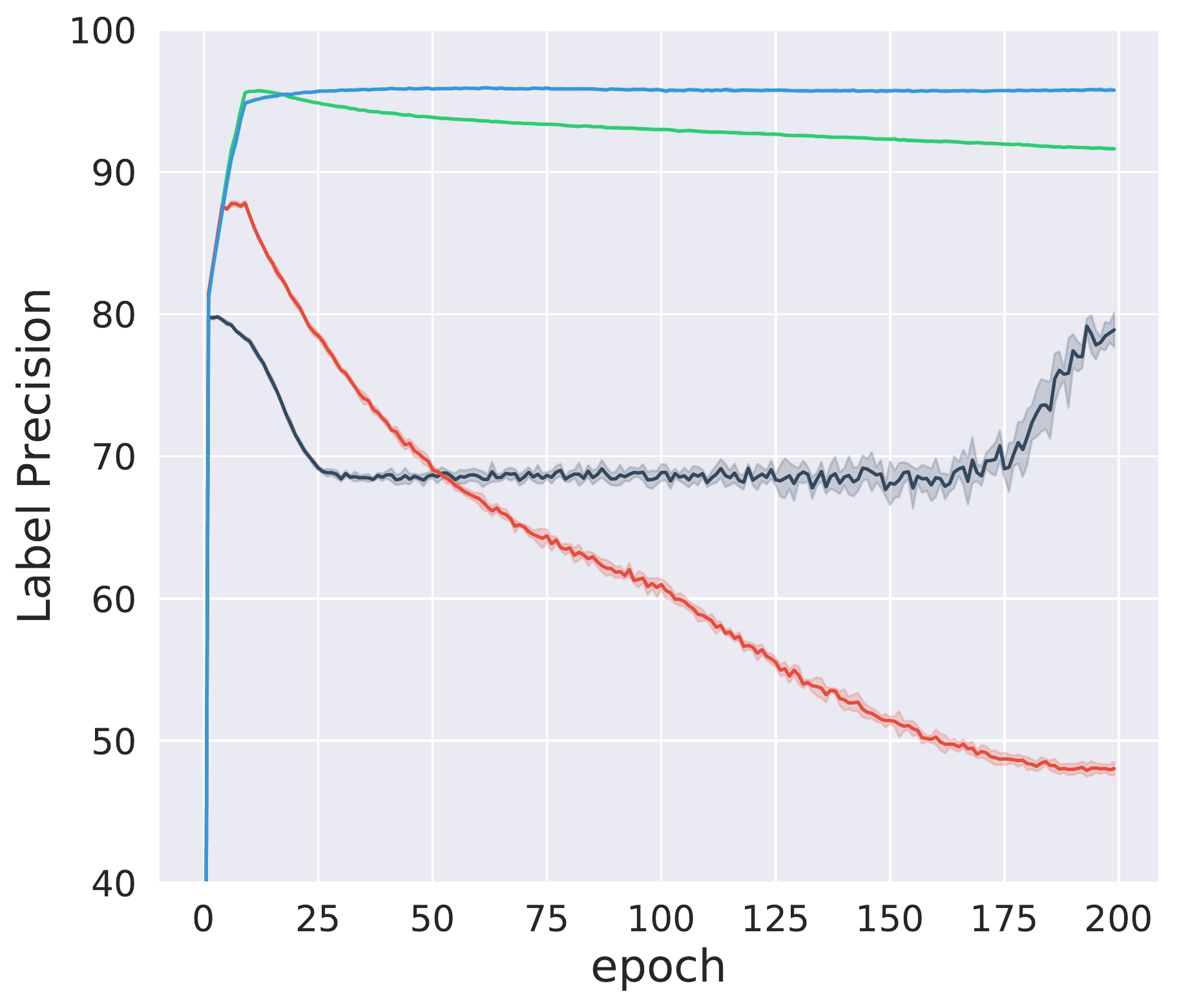}
    \end{minipage}
    }
    \subfigure[Symmetry-50\%]{
    \begin{minipage}[t]{0.2\textwidth}
        \centering
        \includegraphics[scale=0.2]{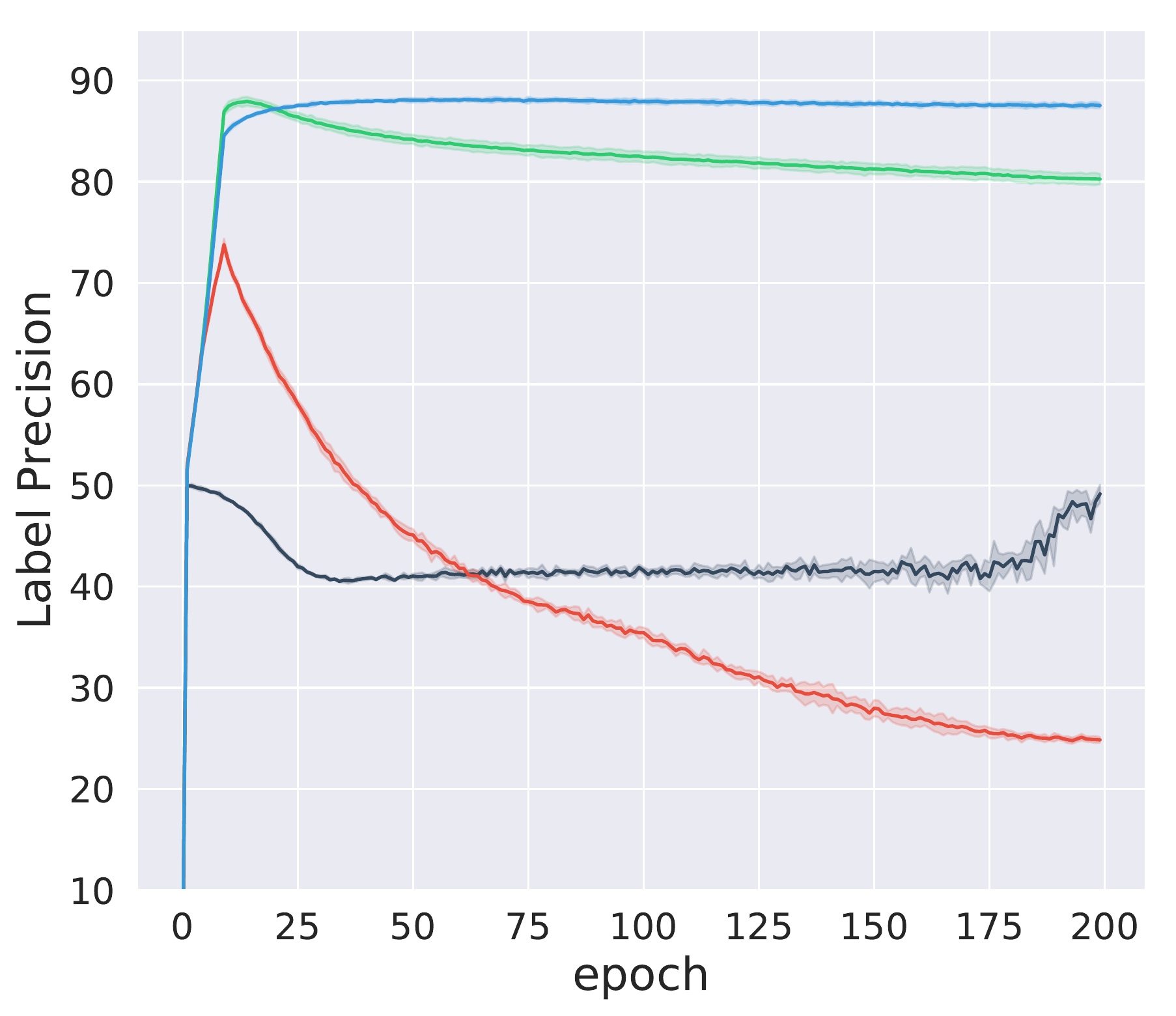}
    \end{minipage}
    }
    \subfigure[Symmetry-80\%]{
    \begin{minipage}[t]{0.2\textwidth}
        \centering
        \includegraphics[scale=0.2]{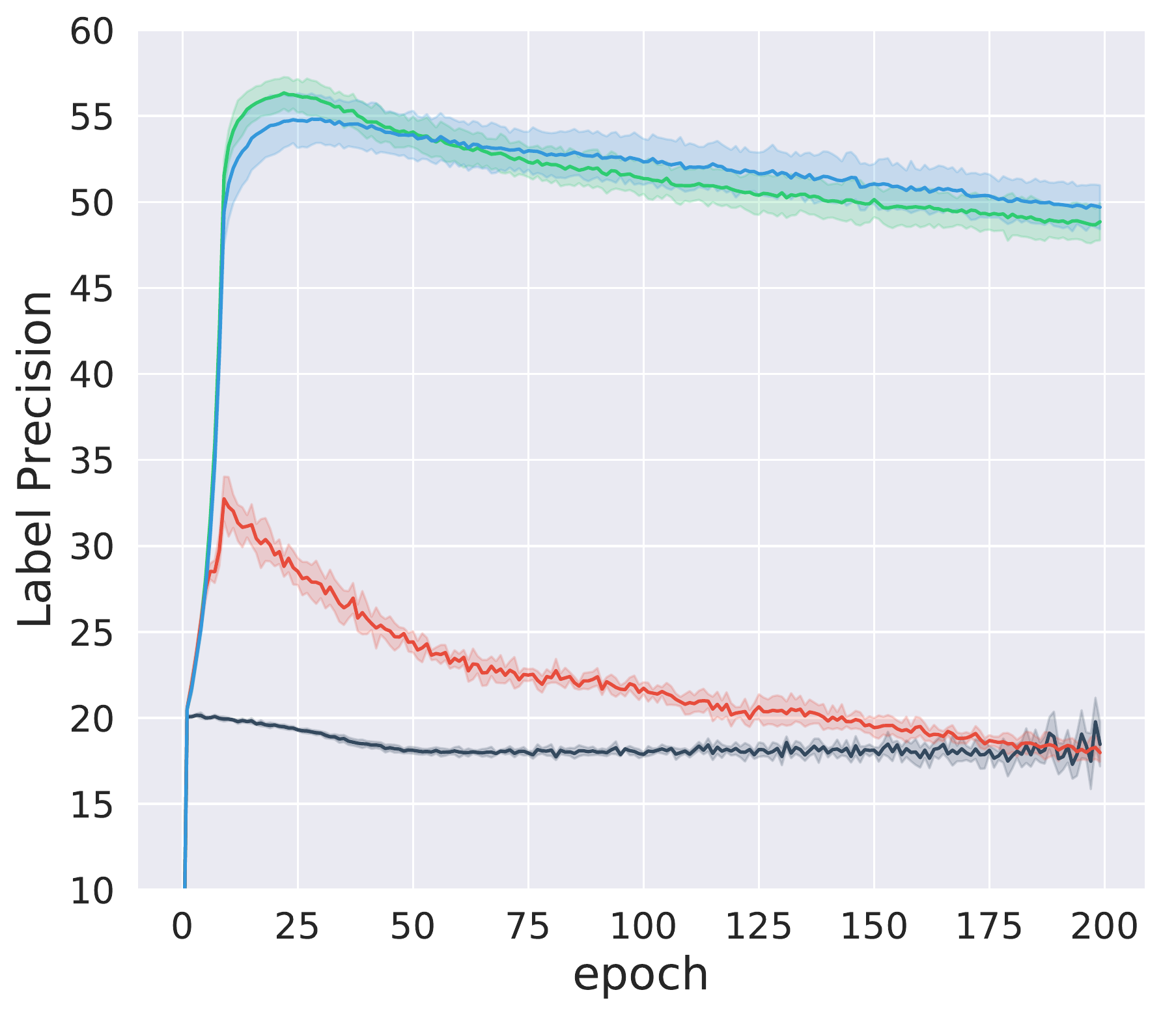}
    \end{minipage}
    }
    \subfigure[Asymmetry-40\%]{
    \begin{minipage}[t]{0.2\textwidth}
        \centering
        \includegraphics[scale=0.2]{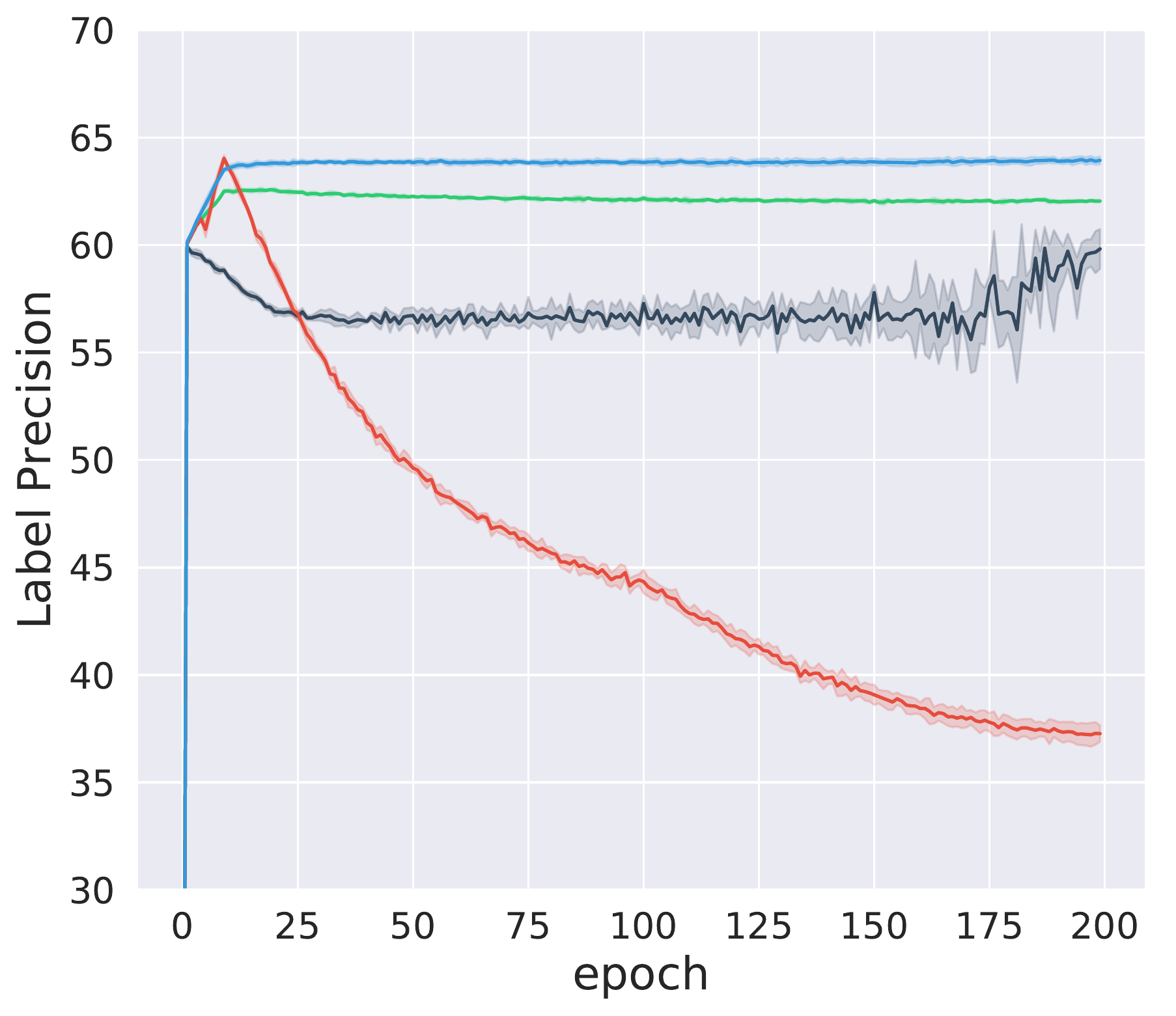}
    \end{minipage}
    }
    \caption{Results on CIFAR-100 dataset. Top: test accuracy(\%) vs. epochs; bottom: label precision(\%) vs. epochs.}
    \vspace{-5pt}
    \label{fig:cifar100_performance}
\end{figure*}

\begin{table*}[ht]
\centering
\topcaption{Average test accuracy (\%) on \textsl{CIFAR-100} over the last 10 epochs.}\label{tab:cifar100_data}
\begin{tabular}{c|c|c|c|c|c|c}

\hline
\hline
Flipping-Rate& Standard & F-correction & Decoupling & Co-teaching & Co-teaching+ & JoCoR\\
\hline
Symmetry-20\% &  $35.14 \pm 0.44$ & $37.95 \pm 0.10$ & $33.10 \pm 0.12$ & $43.73 \pm 0.16$ & $49.27 \pm 0.03$ & $\textbf{53.01} \pm 0.04$ \\
\hline
Symmetry-50\% &  $16.97 \pm 0.40$ & $24.98 \pm 1.82$ & $15.25 \pm 0.20$ & $34.96 \pm 0.50$ & $40.04 \pm 0.70$ & $\textbf{43.49} \pm 0.46$ \\
\hline
Symmetry-80\% &  $4.41 \pm 0.14$ & $2.10 \pm 2.23$ & $3.89 \pm 0.16$ & $15.15 \pm 0.46$ & $13.44 \pm 0.37$ & $\textbf{15.49} \pm 0.98$ \\
\hline
Asymmetry-40\% &  $27.29 \pm 0.25$ & $25.94 \pm 0.44$ & $26.11 \pm 0.39$ & $28.35 \pm 0.25$ & $\textbf{33.62} \pm 0.39$ & $32.70 \pm 0.35$ \\
\hline
\hline
\end{tabular}
\vspace*{-15pt}
\end{table*}

\section{Experiments}

In this section we first compare JoCoR with some state-of-the-art approaches, then analyze the impact of Joint Training and Co-Regularization by ablation study. We also analyze the effect of $\lambda$ in \eqref{eq:joint_loss} by sensitivity analysis and put it in supplementary materials.

\subsection{Experiment setup}

\noindent \textbf{Datasets}. We verify the effectiveness of our proposed algorithm on four benchmark datasets: \textsl{MNIST}, \textsl{CIFAR-10}, \textsl{CIFAR-100} and \textsl{Clothing1M} \cite{xiao2015learning}, and the detailed characteristics of these datasets can be found in supplementary materials. These datasets are popularly used for the evaluation of learning with noisy labels in previous literatures \cite{goldberger2016training,kiryo2017positive,reed2014training}. Especially, \textsl{Clothing1M} is a large-scale real-world dataset with noisy labels, which is widely used in the related works\cite{li2019learning,patrini2017making,yi2019probabilistic,xiao2015learning}.

Since all datasets are clean except \textsl{Clothing1M}, following \cite{patrini2017making,reed2014training}, we need to corrupt these datasets manually by the label transition matrix Q, where $Q_{ij} = \Pr[\tilde{y} = j|y = i]$ given that noisy $\tilde{y}$ is flipped from clean y. Assume that the matrix Q has two representative structures: (1) Symmetry flipping \cite{van2015learning}; (2) Asymmetry flipping \cite{patrini2017making}: simulation of fine-grained classification with noisy labels, where labellers may make mistakes only within very similar classes.

Following F-correction \cite{patrini2017making}, only half of the classes in the dataset are with noisy labels in the setting of asymmetric noise, so the actual noise rate in the whole dataset $\tau$ is half of the noisy rate in the noisy classes. Specifically, when the asymmetric noise rate is 0.4, it means  $\tau=0.2$. Figure \ref{fig:matrix} shows an example of noise transition matrix.

For experiments on \textsl{Clothing1M}, we use the 1M images with noisy labels for training, the 14k and 10k clean data for validation and test, respectively. Note that we do not use the 50k clean training data in all the experiments because only noisy labels are required during the training process \cite{li2019learning,tanaka2018joint}. For preprocessing, we resize the image to $256 \times 256$, crop the middle $224 \times 224$ as input, and perform normalization.

\noindent \textbf{Baselines}. We compare JoCoR (Algorithm \ref{alg:JoCoR}) with the following state-of-the-art algorithms, and implement all methods with default parameters by PyTorch, and conduct all the experiments on NVIDIA Tesla V100 GPU.

\begin{enumerate}[(i)]
\setlength\itemsep{0.005em}
    \item Co-teaching+ \cite{yu2019does}, which trains two deep neural networks and consists of disagreement-update step and cross-update step.
    \item Co-teaching \cite{han2018co}, which trains two networks simultaneously and cross-updates parameters of peer networks.
    \item  Decoupling \cite{malach2017decoupling}, which updates the parameters only using instances which have different predictions from two classifiers.
    \item F-correction \cite{patrini2017making}, which corrects the prediction by the label transition matrix. As suggested by the authors, we first train a standard network to estimate the transition matrix Q.
    \item As a simple baseline, we compare JoCoR with the standard deep network that directly trains on noisy datasets (abbreviated as Standard).
\end{enumerate}

\noindent \textbf{Network Structure and Optimizer}. We use a 2-layer MLP for \textsl{MNIST}, a 7-layer CNN network architecture for \textsl{CIFAR-10} and \textsl{CIFAR-100}. The detailed information can be found in supplementary materials. For \textsl{Clothing1M}, we use ResNet with 18 layers. 

For experiments on \textsl{MNIST}, \textsl{CIFAR-10} and \textsl{CIFAR-100}, Adam optimizer (momentum=0.9) is used with an initial learning rate of 0.001, and the batch size is set to 128. We run 200 epochs in total and linearly decay learning rate to zero from 80 to 200 epochs. 

For experiments on \textsl{Clothing1M}, we also use Adam optimizer (momentum=0.9) and set batch size to 64. During the training stage, we run 15 epochs in total and set learning rate to $8\times10^{-4}$, $5\times10^{-4}$ and $5\times10^{-5}$ for 5 epochs each.

As for $\lambda$ in our loss function \eqref{eq:joint_loss}, we search it in [0.05, 0.10, 0.15,$\ldots$,0.95] with a clean validation set for best performance. When validation set is also with noisy labels, we use the small-loss selection to choose a clean subset for validation. As deep networks are highly nonconvex, even with the same network and optimization method, different initializations can lead to different local optimum. Thus, following Decoupling \cite{malach2017decoupling}, we also take two networks with the same architecture but different initializations as two classifiers.

\noindent \textbf{Measurement}. To measure the performance, we use the test accuracy, i.e., \textit{test accuracy = (\# of correct predictions) / (\# of test)}. Besides, we also use the label precision in each mini-batch, i.e., \textit{label precision = (\# of clean labels) / (\# of all selected labels)}. Specifically, we sample $R(t)$ of small-loss instances in each mini-batch and then calculate the ratio of clean labels in the small-loss instances. Intuitively, higher label precision means less noisy instances in the mini-batch after sample selection, so the algorithm with higher label precision is also more robust to the label noise. All experiments are repeated five times. The error bar for STD in each figure has been highlighted as a shade.

\noindent \textbf{Selection setting}. Following Co-teaching, we assume that the noise rate $\tau$ is known. To conduct a fair comparison in benchmark datasets, we set the ratio of small-loss samples $R(t)$ as identical: 
$R(t) = 1 - \min \left\{ \frac{t}{T_k}\tau ,\tau\right\}$, where $T_k=10$ for MNIST, CIFAR-10 and CIFAR100, $T_k=5$ for Clothing1M. If $\tau$ is not known in advance, $\tau$ can be inferred using validation sets \cite{liu2015classification,yu2018efficient}.

\subsection{Comparison with the State-of-the-Arts}

\noindent \textbf{Results on \textsl{MNIST}}. At the top of Figure \ref{fig:mnist_performance}, it shows test accuracy vs. epochs on \textsl{MNIST}. In all four plots, we can see the memorization effect of networks, i.e., test accuracy of Standard first reaches a very high level and then gradually decreases.  Thus, a good robust training method should stop or alleviate the decreasing process. On this point, JoCoR consistently achieves higher accuracy than all the other baselines in all four cases.

 We can compare the test accuracy of different algorithms in detail in Table \ref{tab:mnist_data}. In the most natural Symmetry-20\% case, all new approaches work better than Standard obviously, which demonstrates their robustness. Among them, JoCoR and Co-teaching+ work significantly better than other methods. When it goes to Symmetry-50\% case and Asymmetry-40\% case, Decoupling begins to fail while other methods still work fine, especially JoCoR and Co-teaching+. However, Co-teaching+ cannot combat with the hardest Symmetry-80\% case, where it only achieves 58.92\%. In this case, JoCoR achieves the best average classification accuracy (84.89\%) again.

To explain such excellent performance, we plot label precision vs. epochs at the bottom of Figure \ref{fig:mnist_performance}. Only Decoupling, Co-teaching, Co-teaching+ and JoCoR are considered here, as they include example selection during training. First, we can see both JoCoR and Co-teaching can successfully pick clean instances out. Note that JoCoR not only reaches high label precision in all four cases but also performs better and better with the increase of epochs while Co-teaching declines gradually after reaching the top. This shows that our approach is better at finding clean instances. Then, Decoupling and Co-teaching+ fail in selecting clean examples. As mentioned in Related Work, very few examples are utilized by Co-teaching+ in the training process when noise rate goes to be extremely high. In this way, we can understand why Co-teaching+ performs poorly on the hardest case.

\noindent \textbf{Results on \textsl{CIFAR-10}}. Table \ref{tab:cifar10_data} shows test accuracy on \textsl{CIFAR-10}. As we can see, JoCoR performs the best in all four cases again. In the Symmetric-20\% case, JoCoR works much better than all other baselines and Co-teaching+ performs better than Co-teaching and Decoupling. In the other three cases, JoCoR is still the best and Co-teaching+ cannot even achieve comparable performance with Co-teaching.

Figure \ref{fig:cifar10_performance} shows test accuracy and label precision vs. epochs. JoCoR outperforms all the other comparing approaches on both test accuracy and label precision. On label precision, while Decoupling and Co-teaching+ fail to find clean instances, both JoCoR and Co-teaching can do this. An interesting phenomenon is that in the Asymmetry-40\% case, although Co-teaching can achieve better performance than JoCoR in the first 100 epochs, JoCoR consistently outperforms it in all the later epochs. The result shows that JoCoR has better generalization ability than Co-teaching.

\noindent \textbf{Results on \textsl{CIFAR-100}}. Then, we show our results on \textsl{CIFAR-100}. The test accuracy is shown in Table \ref{tab:cifar100_data}. Test accuracy and label precision vs. epochs are shown in Figure \ref{fig:cifar100_performance}. Note that there are only 10 classes in \textsl{MNIST} and \textsl{CIFAR-10} datasets. Thus, overall the accuracy is much lower than previous ones in Tables \ref{tab:mnist_data} and \ref{tab:cifar10_data}. But JoCoR still achieves high test accuracy on this datasets. In the easiest Symmetry-20\% and Symmetry-50\% cases, JoCoR works significantly better than Co-teaching+, Co-teaching and other methods. In the hardest Symmetry-80\% case, JoCoR and Co-teaching tie together but JoCoR still gets higher testing accuracy. When it turns to Asymmetry-40\% case, JoCoR and Co-teaching+ perform much better than other methods. On label precision, JoCoR keeps the best performance in all four cases.

\begin{table}[!t]
\centering
\caption{Classification accuracy (\%) on the \textsl{Clothing1M} test set}\label{tab:clothing1m_data}
\begin{tabular}{p{70pt}<{\centering}|p{60pt}<{\centering}|p{60pt}<{\centering}}
\hline
\hline
Methods & \textit{best} & \textit{last}\\
\hline
Standard &  67.22 & 64.68\\
\hline
F-correction &  68.93 & 65.36\\
\hline
Decoupling &  68.48 & 67.32\\
\hline
Co-teaching &  69.21 & 68.51\\
\hline
Co-teaching+ &  59.32 & 58.79\\
\hline
JoCoR &  \textbf{70.30} & \textbf{69.79}\\
\hline
\hline

\end{tabular}
\vspace*{-16pt}

\end{table}

\noindent \textbf{Results on \textsl{Clothing1M}}. Finally, we demonstrate the efficacy of the proposed method on the real-world noisy labels using the \textsl{Clothing1M} dataset. As shown in Table \ref{tab:clothing1m_data}, \textit{best} denotes the scores of the epoch where the validation accuracy is optimal, and \textit{last} denotes the scores at the end of training. The proposed JoCoR method gets better result than state-of-the-art methods on \textit{best}. After all epochs, JoCoR achieves a significant improvement in accuracy of +5.11 over Standard, and an improvement of +1.28 over the best baseline method.

\subsection{Ablation Study}

To conduct ablation study for analyzing the effect of Co-Regularization, we set up the experiments above \textsl{MNIST} and \textsl{CIFAR-10} with Symmetry-50\% noise. For implementing Joint Training without Co-Regularization (\textsl{Joint-only}), we set the $\lambda$ in \eqref{eq:joint_loss} to 0. Besides, to verify the effect of the Joint Training paradigm, we introduce Co-teaching and Standard enhanced by ``small-loss" selection (\textsl{Standard+}) to join the comparison. Recall that the joint-training method selects examples by the joint loss while Co-teaching uses cross-update method to reduce the error flow \cite{han2018co}, these two methods should play a similar role during training according to the previous analysis.

\begin{figure}[!t]
    \centering
    \subfigure{
        \centering
        \includegraphics[scale=0.21]{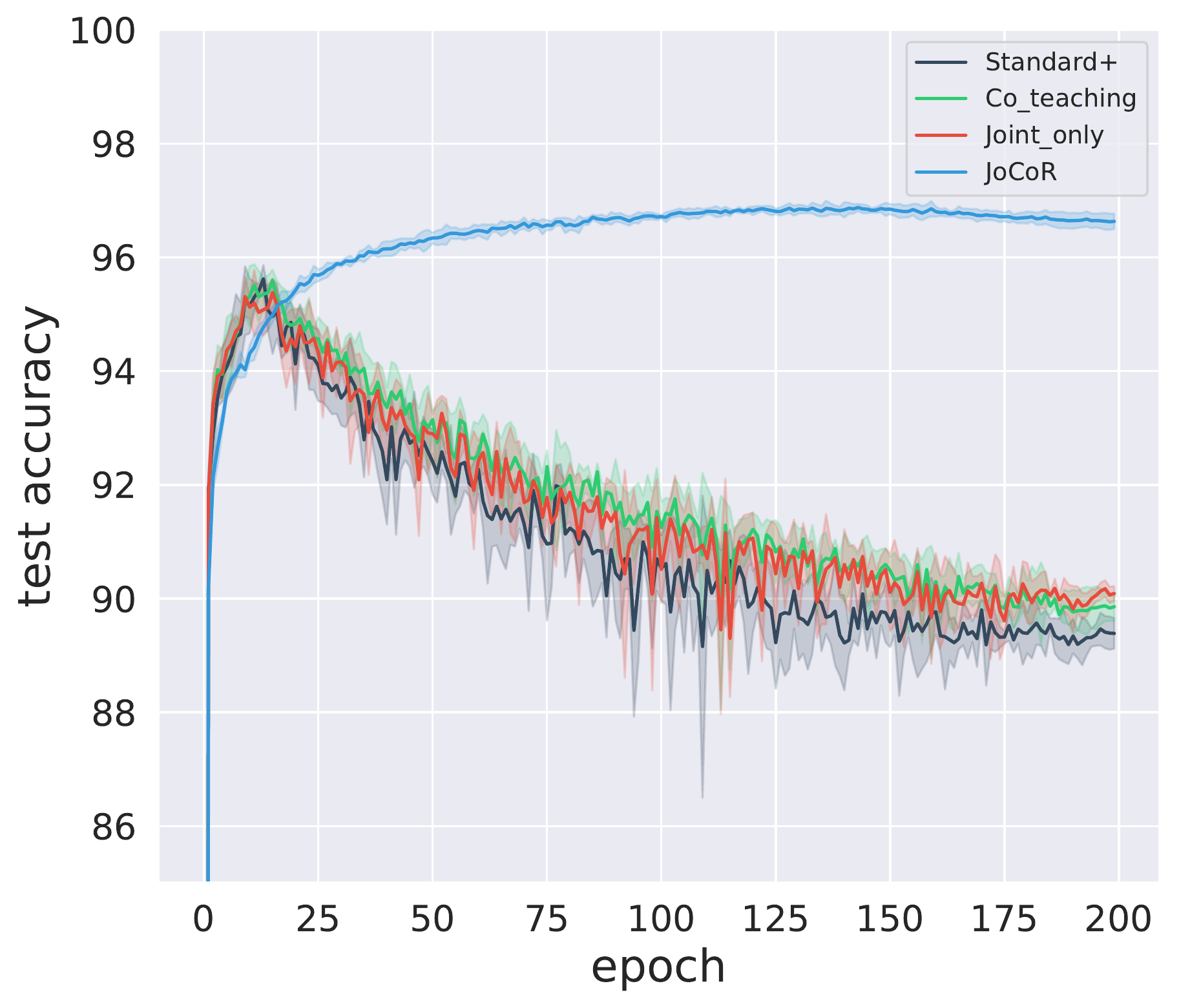}
    }
    \subfigure{
        \centering
        \includegraphics[scale=0.21]{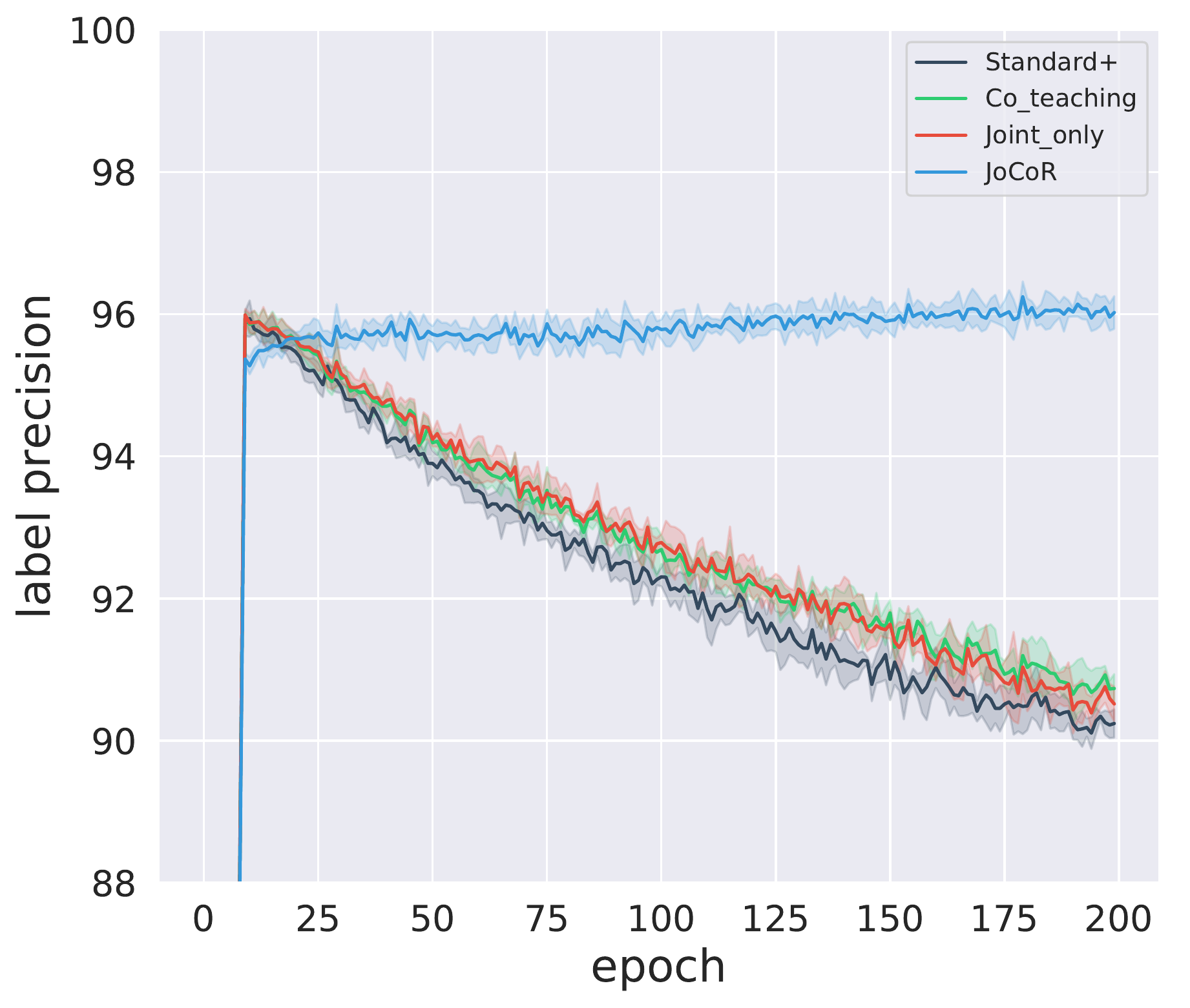}
    }
    \caption{Results of ablation study on \textsl{MNIST}}
    \vspace{-10pt}
    \label{fig:mnist_ablation}
\end{figure}

The test accuracy and label precision vs. epochs on \textsl{MNIST} are shown in Figure \ref{fig:mnist_ablation}. As we can see, JoCoR performs much better than the others on both test accuracy and label precision. The former keeps almost no decrease while the latter decline a lot after reaching the top. This observation indicates that Co-Regularization strongly hinders neural networks from memorizing noisy labels.

\begin{figure}[!t]
    \centering
    \subfigure{
        \centering
        \includegraphics[scale=0.21]{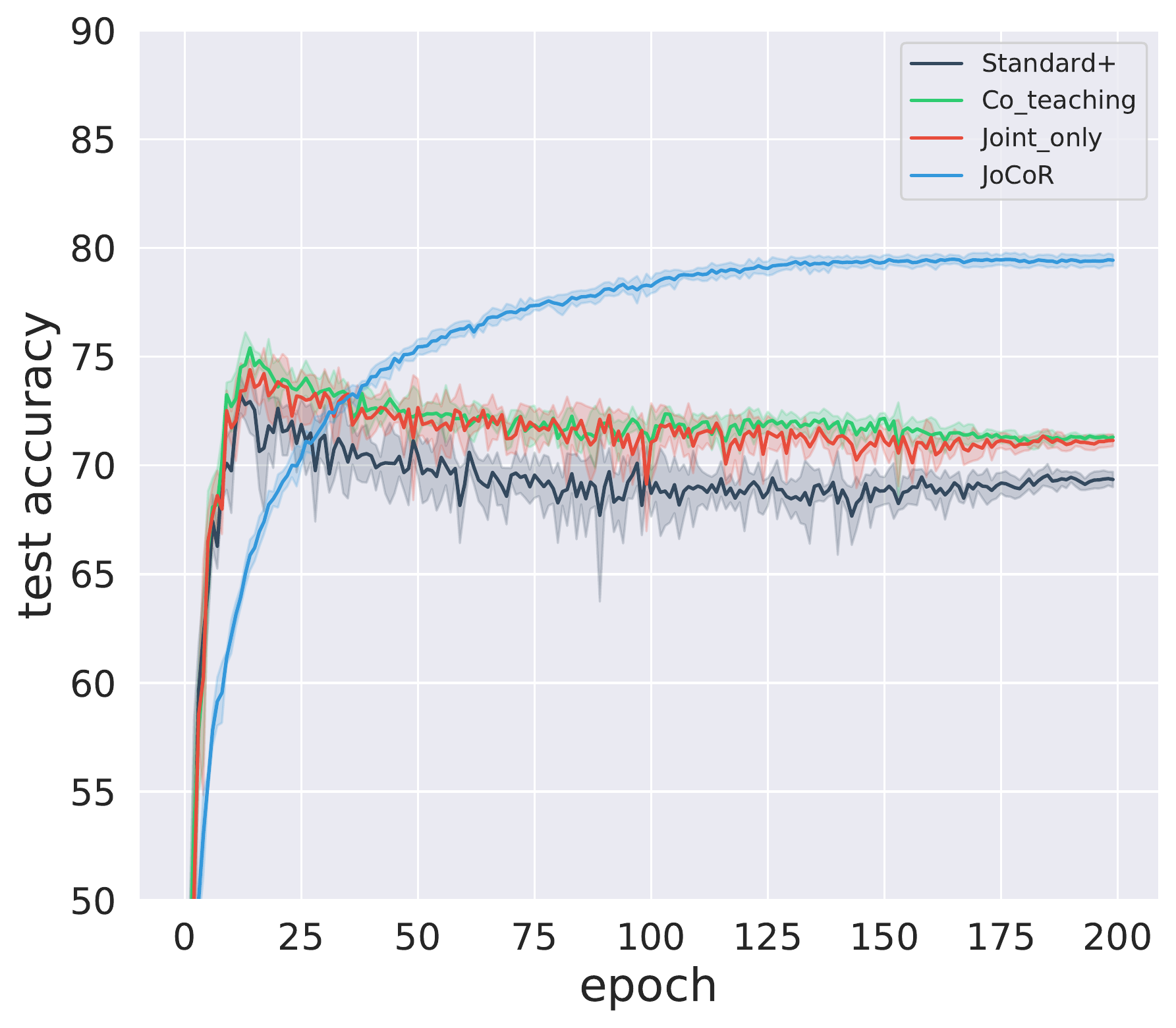}
    }
    \subfigure{
        \centering
        \includegraphics[scale=0.21]{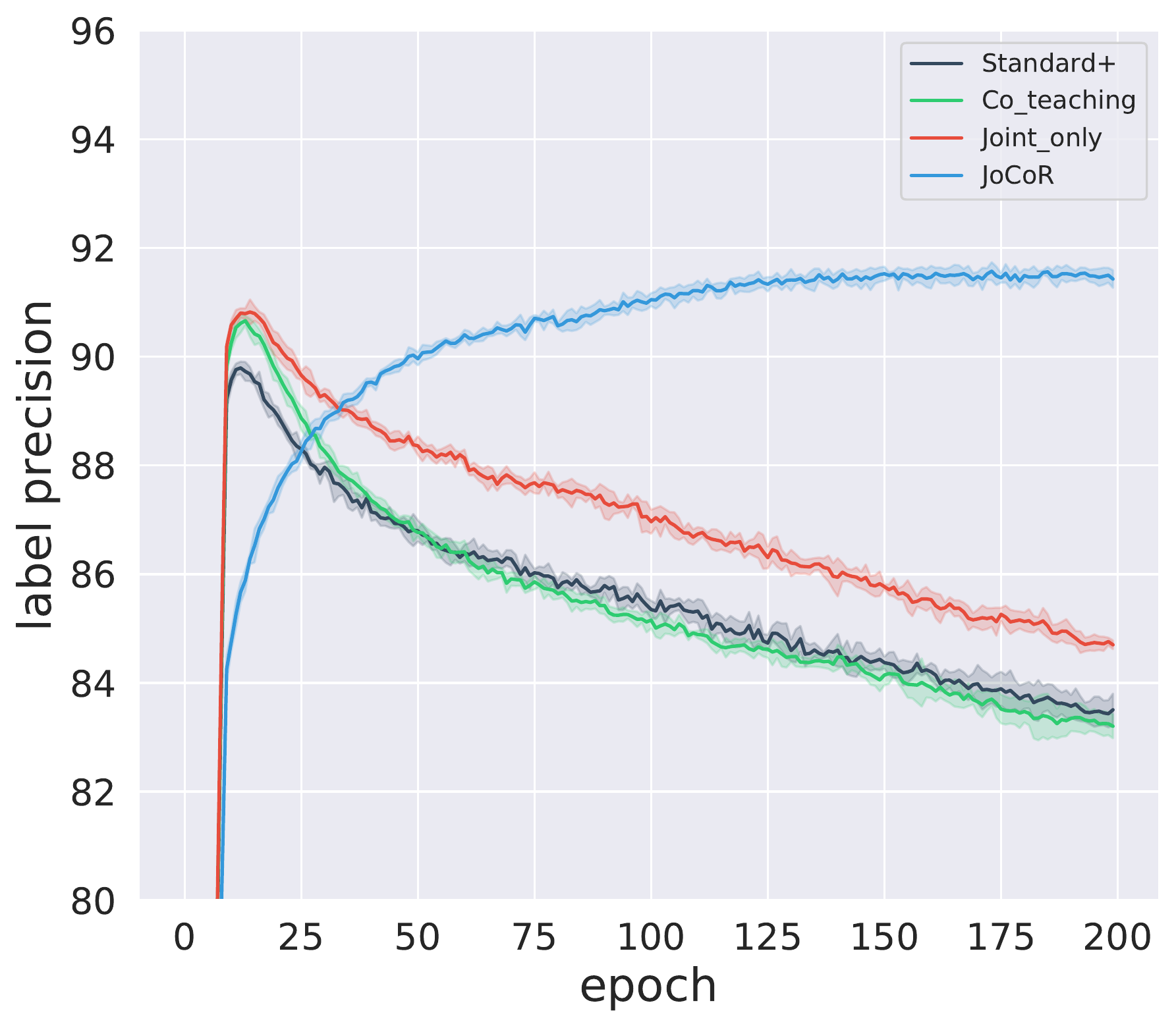}
    }
    \caption{Results of ablation study on \textsl{CIFAR-10}}
    \vspace{-10pt}
    \label{fig:cifar10_ablation}
\end{figure}

The test accuracy and label precision vs. epochs on \textsl{CIFAR-10} are shown in Figure \ref{fig:cifar10_ablation}. In this figure, JoCoR still maintains a huge advantage over the other three methods on both test accuracy and label precision while Joint-only, Co-teaching and Standard+ remain the same trend as these for \textsl{MNIST}, keeping a downward tendency after increasing to the highest point. These results show that Co-Regularization plays a vital role in handling noisy labels. Moreover, Joint-only achieves a comparable performance with Co-teaching on test accuracy and performs better than Co-teaching and Standard+ on label precision. It shows that Joint Training is a more efficient paradigm to help select clean examples than cross-update in Co-teaching.

\section{Conclusion}
The paper proposes an effective approach called JoCoR to improve the robustness of deep neural networks with noisy labels. The key idea of JoCoR is to train two classifiers simultaneously with one joint loss, which is composed of regular supervised part and Co-Regularized part. Similar to Co-teaching+, we also select small-loss instances to update networks in each mini-batch data by the joint loss. We conduct experiments on \textsl{MNIST}, \textsl{CIFAR-10}, \textsl{CIFAR-100} and \textsl{Clothing1M} to demonstrate that, JoCoR can train deep models robustly with the slightly and extremely noisy supervision. Furthermore, the ablation studies clearly demonstrate the effectiveness of Co-Regularization and Joint Training. In future work, we will explore the theoretical foundation of JoCoR based on the view of traditional Co-training algorithms \cite{kumar2010co,sindhwani2005co}.

\noindent \textbf{Acknowledgments} This research is supported by Singapore National Research Foundation projects AISG-RP-2019-0013, NSOE-TSS2019-01, and NTU. We gratefully acknowledge the support of NVAITC (NVIDIA AI Tech Center) for our research.

{\small
\bibliographystyle{ieee_fullname}
\bibliography{egbib}
}

\clearpage

\begin{appendix}
\section{Dataset}

the detailed characteristics of these datasets are shown in Table \ref{tab:dataset}.

\begin{table}[ht]\footnotesize
\centering
\topcaption{Summary of datasets used in the experiments.}\label{tab:dataset}

\begin{tabular}{c|c|c|c|c}
\hline
\hline
 & \# of training & \# of test & \# of class & size \\
\hline
\textsl{MNIST} &  60,000 & 10,000 & 10 & $28 \times 28$ \\
\hline
\textsl{CIFAR-10} &  50,000 & 10,000 & 10 & $32 \times 32$ \\
\hline
\textsl{CIFAR-100} &  50,000 & 10,000 & 100 & $32 \times 32$ \\
\hline
\textsl{Clothing1M} &  1,000,000 & 10,000 & 14 & $224 \times 224$ \\
\hline
\hline
\end{tabular}
\end{table}

\section{Network Architecture}

The network architectures of the MLP and CNN models are shown in Table \ref{tab:model}.

\begin{table}[ht]

\centering
\footnotesize
%\top
\caption{The models used on \textsl{MNIST}, \textsl{CIFAR-10} and \textsl{CIFAR-100}}\label{tab:model}
\begin{tabular}{p{3.5cm}<{\centering}|p{3.5cm}<{\centering}}
\hline
\hline
MLP on \textsl{MNIST}& CNN on \textsl{CIFAR-10}\& \textsl{CIFAR-100} \\
\hline
$28 \times 28$ Gray Image& $32 \times 32$ RGB Image\\
\hline
\multirow{6}*{Dense $28 \times 28 \longrightarrow 256$, ReLU}& \tabincell{c}{$3 \times 3$, 64 BN, ReLU \\ $3 \times 3$, 64 BN, ReLU \\ $2 \times 2$ Max-pool} \\ 
\cline{2-2} & \tabincell{c}{$3 \times 3$, 128 BN, ReLU \\ $3 \times 3$, 128 BN, ReLU \\ $2 \times 2$ Max-pool} \\ 
\cline{2-2} & \tabincell{c}{$3 \times 3$, 196 BN, ReLU \\ $3 \times 3$, 196 BN, ReLU \\ $2 \times 2$ Max-pool} \\
\hline
Dense $ 256 \longrightarrow 10$ & Dense $ 256 \longrightarrow 100$ \\
\hline
\hline
\end{tabular}
\end{table}

\section{Parameter Sensitivity Analysis}

To conduct the sensitivity analysis on parameter $\lambda$, we set up the experiments above \textsl{MNIST} with symmetry-50\% noise. Specifically, we compare $\lambda$ in the range of [0.05, 0.35, 0.65, 0.95]. The larger the $\lambda$ is, the less the divergence of two classifiers in JoCoR would be.

\begin{figure}[ht]
    \centering
    \subfigure[test accuracy]{
        \centering
        \includegraphics[scale=0.19]{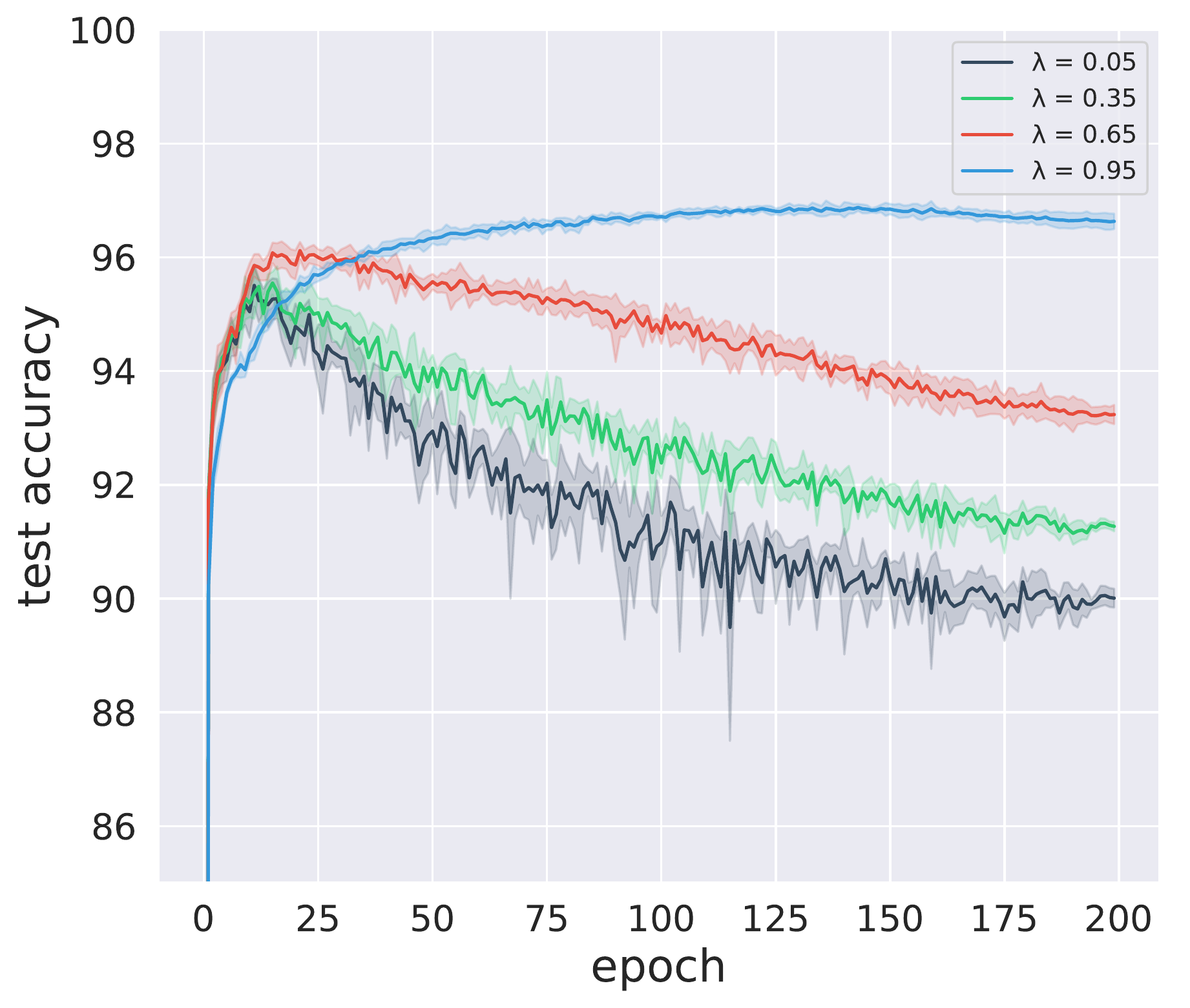}
    }
    \subfigure[label precision]{
        \centering
        \includegraphics[scale=0.19]{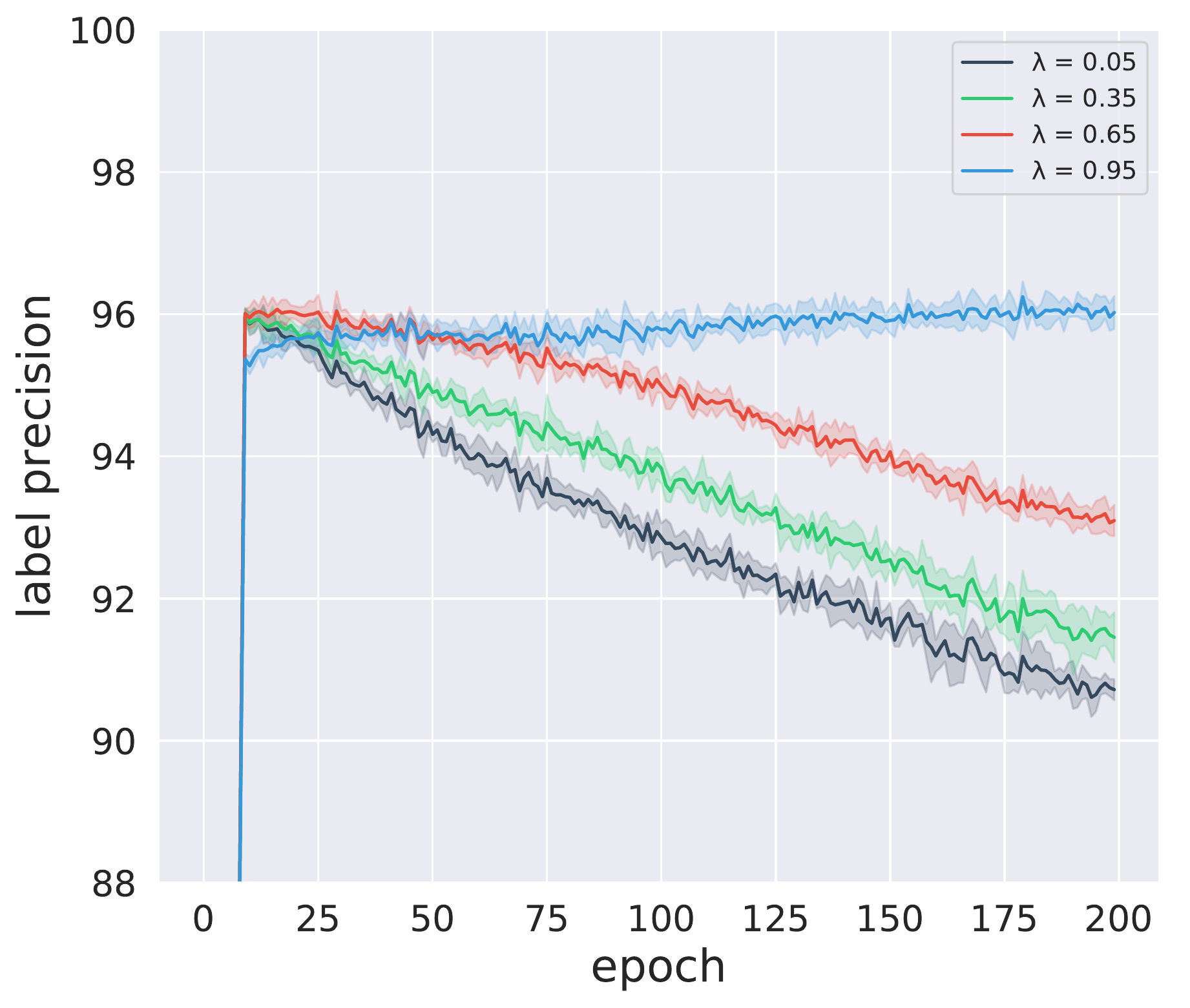}
    }
    \caption{Results of JoCoR with different $\lambda$ on \textsl{MNIST}}
    \label{fig:psa}
\end{figure}

The test accuracy and label precision vs. number of epochs are in Figure \ref{fig:psa}. Obviously, As the $\lambda$ increases, the performance of our algorithm on test accuracy gets better and better. When $\lambda = 0.95$, JoCoR achieves the best performance. We can see the same trends on label precision, which means that JoCoR can select clean example more precisely with a larger $\lambda$.

\end{appendix}

\end{document}